\documentclass[10pt,twocolumn,letterpaper]{article}
\usepackage{iccv}
\usepackage{times}
\usepackage{epsfig}
\usepackage{graphicx}
\usepackage{amsmath}
\usepackage{amssymb}
\usepackage{marvosym}
\usepackage{lipsum}
\usepackage{float}
\usepackage{multirow}
\usepackage{subfigure}
\usepackage{tikz,pgfplots}
\usepackage{amsmath}
\usepackage{amssymb}
\usepackage{booktabs}
\usepackage{graphicx}
\usepackage{multirow}
\usepackage{nccfoots}
\usepackage{tabularx}
\usepackage{bbm}
\usepackage{comment}
\usepackage{threeparttable}
\usepackage{enumitem}
\usepackage{caption}
\usepackage{color, colortbl}
\usepackage{stfloats}
\usepackage{pifont}
\usepackage{footnote}

\newcommand{\cmark}{\text{\ding{51}}} 
\newcommand{\xmark}{\text{\ding{55}}} 

\definecolor{citecolor}{HTML}{0071bc}
\usepackage[pagebackref=false,breaklinks=true,letterpaper=true,colorlinks,citecolor=citecolor,bookmarks=false]{hyperref}
\usepackage[capitalize]{cleveref}
\crefname{section}{Sec.}{Secs.}
\Crefname{section}{Section}{Sections}
\Crefname{table}{Table}{Tables}
\crefname{table}{Tab.}{Tabs.}

\definecolor{citecolor}{HTML}{0071bc}
\definecolor{color_ao}{gray}{0.5}
\definecolor{color_our}{rgb}{0.66,0.82,0.56}
\definecolor{color_pre}{rgb}{0.52,0.59,0.69}
\definecolor{Gray}{gray}{0.9}
\definecolor{LighterGray}{gray}{0.93}
\definecolor{LightGrayForTableRule}{gray}{0.92}
\definecolor{DarkGray}{gray}{0.5}
\definecolor{Black}{rgb}{0.0, 0.0, 0.0}
\definecolor{NiceBlue}{rgb}{0.11764705882352941, 0.5647058823529412, 1.0}
\definecolor{NiceGreen}{rgb}{0.0, 0.5, 0.0}
\usepackage{pgfkeys}
	\def\addlegendimage{\csname pgfplots@addlegendimage\endcsname}

\iccvfinalcopy %
\def\iccvPaperID{3426} 

\definecolor{Yellow}{RGB}{255, 242, 204}
\definecolor{Cyan}{RGB}{202, 242, 204}

\begin{document}

\newcommand{\vtg}{VTG}
\newcommand{\vlp}{VLP}
\newcommand{\VTG}{Video Temporal Grounding}
\newcommand{\VLP}{Vision-Language Pretraining}
\newcommand{\VLTG}{Video-Language Temporal Grounding}
\newcommand{\our}{{UniVTG}}
\newcommand{\unit}{clip}

\newcommand{\mr}{moment retrieval}
\newcommand{\hl}{highlight detection}
\newcommand{\vsum}{video summarization}
\newcommand{\tal}{temporal action localization}

\newcommand{\MR}{Moment Retrieval}
\newcommand{\HL}{Highlight Detection}
\newcommand{\VS}{Video Summarization}
\newcommand{\TAL}{Temporal Action Localization}

\newcommand{\data}{label}
\newcommand{\Interval}{Interval-wise}
\newcommand{\Curve}{Curve-wise}
\newcommand{\Point}{Point-wise}
\newcommand{\interval}{interval}
\newcommand{\curve}{curve}
\newcommand{\point}{point}
\newcommand{\intervall}{interval-level}
\newcommand{\curvel}{curve-level}
\newcommand{\pointl}{point-level}

\newcommand{\format}{unified \data~format}
\newcommand{\Format}{Unified \data~format}

\newcommand{\ind}{foreground indicator}
\newcommand{\bdy}{boundary offsets}
\newcommand{\sal}{saliency score}

\newcommand{\detr}{Moment-DETR}
\newcommand{\iv}{IntentVizor}

\newcommand{\qv}{QVHighlights}
\newcommand{\charades}{Charades-STA}
\newcommand{\anet}{Activitynet Captions}
\newcommand{\ego}{Ego4D}
\newcommand{\youtube}{YouTube Highlights}
\newcommand{\tvsum}{TVSum}
\newcommand{\qfvs}{QFVS}
\newcommand{\tacos}{TACoS}
\newcommand{\nlq}{NLQ}
\newcommand{\NLQ}{Natural Language Queries}
\newcommand{\supp}{Supplementary}
\newcommand{\sota}{state-of-the-art}

\title{\our: Towards Unified \VLTG}

\author{Kevin Qinghong Lin$^1$,
~ Pengchuan Zhang$^2$,~
Joya Chen$^1$,
Shraman Pramanick$^3$,\\
Difei Gao$^1$,~
Alex Jinpeng Wang$^1$,~
Rui Yan$^1$,
Mike Zheng Shou$^1$\textsuperscript{\Letter}
\vspace{1.5mm}
\\
\vspace{-2mm}
$^1$Show Lab, National University of Singapore\quad $^2$Meta AI\quad $^3$Johns Hopkins University\\
}

\maketitle
\newcommand\blfootnote[1]{%
  \begingroup
  \renewcommand\thefootnote{}\footnote{#1}%
  \addtocounter{footnote}{-1}%
  \endgroup
}

\ificcvfinal

\begin{abstract}
\vspace{-0.5em}
\VTG~(\vtg), which aims to ground target \unit s from videos (such as consecutive intervals or disjoint shots) according to custom language queries (e.g., sentences or words), 
is key for video browsing on social media.
Most methods in this direction develop task-specific models that are trained with type-specific \data s, 
such as \mr~(time~\interval) and \hl~(worthiness \curve),
which limits their abilities to generalize to various \vtg~tasks and labels.
In this paper, we propose to \underline{Uni}fy the diverse \underline{\vtg} \data s and tasks, dubbed \textup{\our}, along three directions:
Firstly, we revisit a wide range of \vtg~labels and tasks and define a unified formulation. Based on this, we develop data annotation schemes to create scalable pseudo supervision.
Secondly, we develop an effective and flexible grounding model capable of addressing each task and 
making full use of each label.
Lastly, thanks to the unified framework, we are able to unlock {temporal grounding pretraining} from large-scale diverse labels and develop stronger grounding abilities e.g., zero-shot grounding.
Extensive experiments on three tasks (\mr, \hl~and \vsum) across seven datasets  (\qv, \charades, \tacos, \ego, \youtube, \tvsum, and \qfvs) 
demonstrate the effectiveness and flexibility of our proposed framework. 
The codes are available at \textcolor{citecolor}{\url{https://github.com/showlab/UniVTG}}.
\end{abstract}
\blfootnote{\Letter: Corresponding Author.}

\vspace{-1.8em}
\section{Introduction}
\vspace{-0.2em}
With the increasing interest in sharing daily lives, video has emerged as the most informative yet diverse visual form on social media.
These videos are collected in a variety of settings,
including untrimmed instructional videos~\cite{miech2019howto100m}, and well-edited vlogs~\cite{lei2021detecting}.
With massive scales and diverse video forms, automatically identifying relevant moments based on user queries has become a critical capability in the industry for efficient video browsing.

\begin{figure}[t]
    \includegraphics[width=1.0\linewidth]{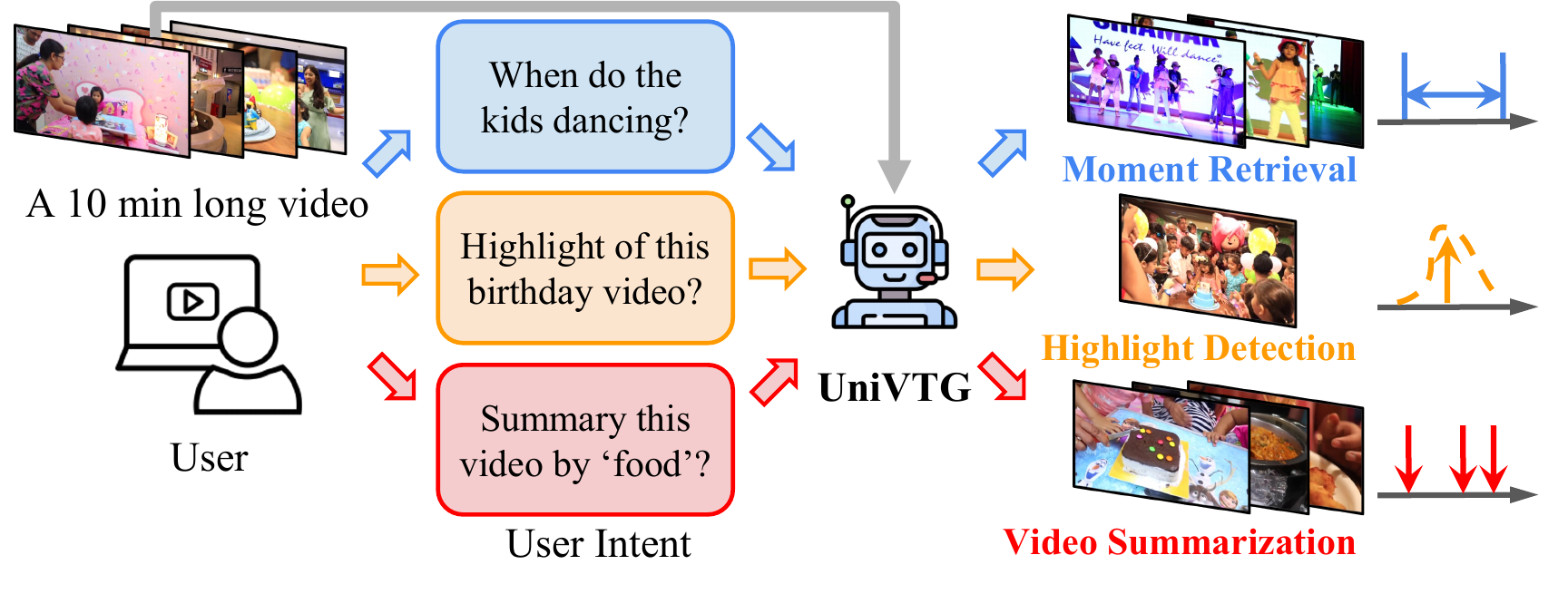}
    \captionsetup{font={small}}
    \vspace{-2.0em}
    \caption{\small{Given a video and a specific user query, \our~serves as a general video browsing helper that assists users by returning different scale target clips 
    to support various \vtg~tasks.
    \vspace{-1.2em}
    }}
    \vspace{-1.2em}
    \label{fig:demo1}
\end{figure}


This significant demand has given rise to a number of video understanding tasks, including \mr~\cite{zhang2020learning, zhang2020span, mun2020local}, \hl~\cite{xiong2019less, hong2020mini, xu2021cross}, and \vsum~\cite{gygli2014creating, song2015tvsum, sharghi2017query}.
As depicted in Fig.~\ref{fig:demo1},
\mr~tends to localize consecutive temporal windows~(\intervall) by giving natural sentences; 
\hl~aims to pick out the key segment with highest worthiness~(\curvel) that best reflects the video gist;
\vsum~collects a set of disjoint shots~(\pointl) to summarize the video, with general or user-specific queries. 
Despite task-specific datasets~\cite{gao2017tall, caba2015activitynet, sun2014ranking, song2015tvsum} and models~\cite{zhang2020learning, zhang2020span, xu2021cross} have been developed, these tasks are typically studied separately.
In general, these tasks share a common objective of grounding various scale \unit s based on customized user queries, which we refer to as {{\VTG}~(\vtg)}. 

Though these tasks are closely related, their relationship has not been explicitly studied until recently.
\cite{lei2020tvr} introduces the first unified benchmark \qv~for \mr~and \hl, and presents the first model \detr~for jointly learning. 
On this basis, UMT~\cite{liu2022umt} expands audio inputs, and QD-DETR~\cite{moon2023query} develops negative-pairs and saliency tokens.
Nevertheless, these studies solely focus on designing models that intersect two subtasks and learn grounding capabilities rely on specific labels. 
This means that they lack the ability to generalize the \vtg~across diverse temporal labels,
such as unique \pointl~narrations in \ego~\cite{grauman2022ego4d}.
Furthermore, we have witnessed promising progress in Vision-Language Pretraining~(\vlp). One notable work is GLIP~\cite{li2022grounded, zhang2022glipv2}, which develops a unified model via joint utilizing large-scale diverse image annotations such as image captions and bounding boxes for spatial grounding.
However, we do not observe similar progress in video-language pretraining. Most works in this area are designed for video-level tasks such as video-text retrieval~\cite{xu2016msr,wang2023all} rather than temporal grounding.
This is largely due to the manual cost of fine-grained temporal annotations is expensive, making it challenging to obtain open-source, scalable yet diverse annotations to support grounding pretraining along the temporal axis in videos.

Therefore, we see a clear motivation to pursue a {Uni}fied {\vtg}~framework and propose our {\our}, which aims
to unify diversity in \vtg~along three directions:
\textbf{(i) From the label and task aspect}, 
we first define a formulation for \vtg~where each video is decomposed as a \unit~sequence that each \unit~is assigned three basic query-conditional elements.
Such a formulation enables us to unify various \vtg~\data s and tasks under the same framework.
Moreover, to address the limitation of temporal labels, we propose a data annotation scheme based on CLIP~\cite{radford2021learning} to produce scalable fine-grained pseudo labels.
\textbf{(ii) From the model aspect}, we develop a flexible yet effective grounding model that inherits the principles of our formulation.
Our model devises single-stream and dual-stream pathways for modality fusion and modality alignment respectively, and is equipped with three heads to decode three key elements.
This favorable design is capable of addressing each task and utilizing each label.
\textbf{(iii)} 
Lastly, thanks to the unified framework and the availability of pseudo labels, we can perform \textbf{large-scale temporal grounding pretraining} across various labels to enhance our grounding abilities. This empowers us to address  various \vtg~downstream tasks across multiple domains, including zero-shot inference.

To validate the effectiveness of our proposed framework, 
we conduct experiments not only on joint \mr~and \hl~benchmark (\qv~\cite{lei2020tvr}), but also on three individual tasks for 
\mr~(\ego~\cite{grauman2022ego4d}, \charades~\cite{gao2017tall}, \tacos~\cite{regneri2013grounding}), \hl~(\youtube~\cite{sun2014ranking}, \tvsum~\cite{song2015tvsum}) and 
\vsum~(QFVS~\cite{sharghi2017query}).
Our \our, one unified model with $4.2$M samples for temporal grounding pretraining, has achieved remarkable results, outperforming \sota~methods that are specifically tailored for each task.
Overall, our contributions are four folds:
\begin{itemize}[itemsep=-2.pt, topsep=0pt]    
    \item To the best of our knowledge, our \our~is the first video temporal grounding pretraining that across varied domains and tasks, including \mr, \hl~and \vsum.
    \item We introduce a unified \vtg~framework that can fully leverage rich supervision from open-source, scalable yet diverse temporal annotations, such as \pointl, \intervall, and \curvel~labels.
    \item To address the limitations of pretraining corpus, 
    we develop an efficient annotation method that uses CLIP as a teacher to produce scalable pseudo temporal labels.
    \item We demonstrate the effectiveness and flexibility of the proposed framework across four settings and seven datasets.
    Detailed ablation studies validate the superiority of the proposed components.
\end{itemize}
\begin{figure}[t]
    \includegraphics[width=1.0\linewidth]{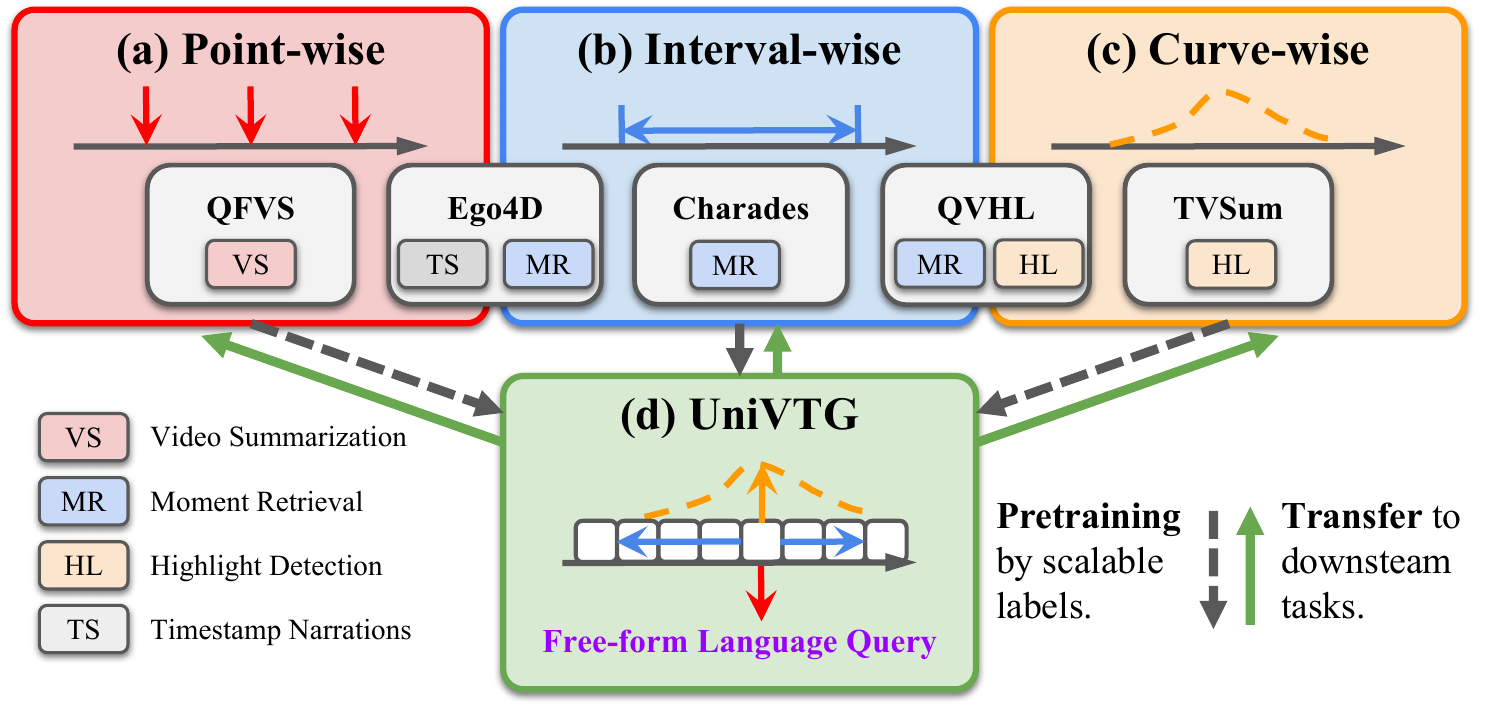}
    \captionsetup{font={small}}
    \caption{\small{Diverse \vtg~\data s can be divided into three types, each mainly
    associated with specific benchmarks:
    \textbf{(a) \pointl~\data s} for \vsum~\cite{sharghi2017query}~and timestamp narrations~\cite{grauman2022ego4d}; 
    \textbf{(b) \intervall~\data s} for \mr~\cite{grauman2022ego4d, gao2017tall, lei2020tvr};
    \textbf{(c) \curvel~\data s} for \hl~\cite{song2015tvsum, lei2020tvr}.
    \textbf{(d) \our}~unifies diverse labels and tasks within one framework, enabling large-scale pretraining with diverse labels (\textit{dotted gray line}) that can be transferred to various downstream tasks (\textit{solid green line}).
    \vspace{-2em}
    }}
    \label{fig:demo2}
\end{figure}

\begin{figure*}[ht]
    \centering
    \vspace{-2mm}
    \includegraphics[width=0.95\linewidth]{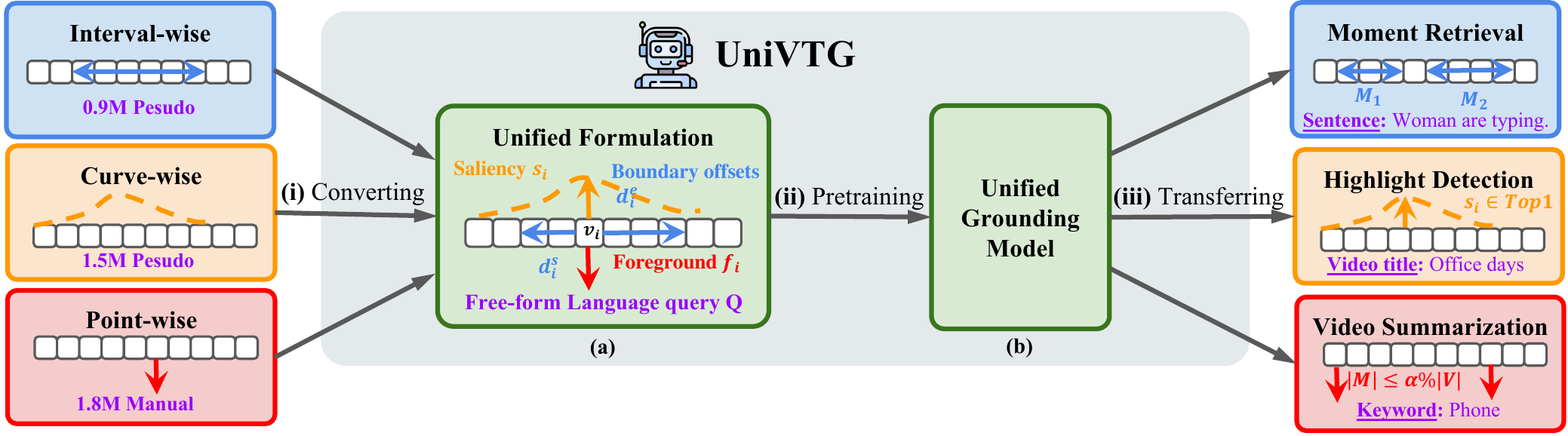}
    \vspace{-2mm}
    \captionsetup{font={small}}
    \caption{\small{
    \textbf{Illustration of \our~pipeline.}
    \textbf{(i)} Given any kind of labels, such as interval label, we first  convert it into our \textbf{(a) unified formulation} ($\S$~\ref{sec:definition}) by deriving other two labels (point and curve labels). 
    \textbf{(ii)} Once we have collect a large-scale diverse labels ($\S$~\ref{sec:corpus}), we leverage them to pretrain a \textbf{unified grounding model} ($\S$~\ref{sec:model}). 
    \textbf{(iii)} Next, the unified model is transferred to various \vtg~downsteam tasks \textit{e.g.,} \hl. 
    \vspace{-1.5em}
    }}
    \label{fig:pipeline}
\end{figure*}

\section{Related Work}
\vspace{-.5em}
\subsection{\VTG}
\vspace{-.5em}
We review three {\vtg} tasks: \mr, \hl, and \vsum, and compare them as different variations of a common problem.

\noindent\textbf{\MR} aims to localize target moments \textit{i.e.,} one~\cite{gao2017tall} or many~\cite{lei2020tvr} continuous intervals within a video by a language query, as shown in Fig.~\ref{fig:demo2}~(b). 
Previous methods fall into two categories: proposal-based and proposal-free.
The proposal-based methods~\cite{anne2017localizing, gao2017tall, zhang2020learning} employ a two-stage process of scanning the entire video to generate candidate proposals, which are then ranked based on their matching to the text query. 
In contrast, the proposal-free methods~\cite{chen2018temporally, zeng2020dense, ghosh2019excl, zhang2020span, mun2020local} learn to regress the start and end boundaries directly without requiring proposal candidates.
Our \our~borrows from proposal-free approaches but extends it by incorporating diverse temporal labels and tasks with a concise design.

\noindent\textbf{Highlight Detection} aims to assign a worthiness score to each video segment \textit{e.g.,} Fig.~\ref{fig:demo2}~(c), and then return the top highest scoring segment as the highlight.
Previous \hl~datasets~\cite{rui2000automatically, sun2014ranking, song2015tvsum} tend to be domain-specific and query-agnostic, in which many efforts~\cite{gygli2016video2gif, xiong2019less, hong2020mini, xu2021cross, badamdorj2021joint} treat this task as a visual or visual-audio scoring problem. 
Nevertheless, video highlights typically have a theme, which is often reflected in the video titles~\cite{song2015tvsum} or topics~\cite{sun2014ranking} \textit{e.g.,} ``surfing''.
Recently, \cite{lei2020tvr}~proposes a joint \mr~and \hl~benchmark \qv~that 
enables users to produce various highlights for one video conditional on different text queries.

\noindent\textbf{Video Summarization} aims to summarize the whole video by a set of shots to provide a quick overview \textit{e.g.,} Fig.\ref{fig:demo2}~(a), which contains two forms: 
Generic \vsum~\cite{gygli2014creating, song2015tvsum, mahasseni2017unsupervised, jiang2022joint} that captures the important scene using visual clues merely, while Query-focused \vsum~\cite{sharghi2017query, nalla2020watch, wu2022intentvizor} that allows users to customize the summary by specifying text keywords (\textit{e.g.,} tree and cars). The latter is closer to practical usage hence we focus on it. Recently, \iv~\cite{wu2022intentvizor} proposes an interactive approach allowing users to adjust their intents to obtain a superior summary.

In general, each of the three tasks represents a specific form of \vtg~that grounds different scales of \unit s from videos (\textit{e.g.,} a consecutive \unit~set, a single \unit~or a disjoint \unit~set) by offering customized text queries (\textit{e.g.,} sentences, titles or keywords). 
However, previous methods address some subtasks solely.
Based on this insight, our goal is to develop a unified framework to handle all of them.

\vspace{-.5em}
\subsection{\VLP}
\vspace{-.5em}
The emergence of large-scale vision-language datasets, such as~\cite{sharma2018conceptual, shao2019objects365, miech2019howto100m, bain2021frozen}, has paved the way for the development of \vlp~\cite{radford2021learning,li2021align,lei2021less,pramanick2022volta,li2022blip} to enhance video-text representation for various vision-language tasks~\cite{young2014image, xu2016msr, msrvttqamsvdqa}.
The representative CLIP~\cite{radford2021learning} has shown that \textit{image-level} visual representations can be effectively learned using large-scale noisy image-text pairs.
Furthermore,  GLIP~\cite{li2022grounded, zhang2022glipv2} makes an effort along the spatial axis, which leverages various image annotations, such as image labels, captions, and bounding boxes, to develop strong \textit{region-level} understanding capacity for spatial grounding tasks.
However, due to the expensive manual cost of fine-grained \textit{temporal-level} annotations \textit{i.e.,} temporal bounding box, this grounding pretraining has not been extended to the temporal axis in videos, limiting its progress to match the spatial counterparts. 
To address this limitation, we explore alternative approaches that leverage accessible timestamp narrations~\cite{grauman2022ego4d} and derive pseudo supervision as the pretraining corpus.

On the other hand, there are several efforts have been made to perform temporal-friendly video pretraining~\cite{alwassel2021tsp, xu2021boundary, cao2022locvtp, zhang2022unsupervised} to pursue a better video representation for grounding tasks.
But the resulting pretraining model still requires an additional grounding model such as 2D-TAN~\cite{zhang2020learning} to perform video grounding.
In contrast, powered by our unified framework and scalable pseudo annotations, we can directly conduct \vlp~with grounding as a pretraining task. This way eliminates the need for additional grounding models and enables zero-shot grounding capacity.
\vspace{-.5em}
\section{Towards Unified VTG: Tasks and Labels}
The \our~pipeline is displayed in Fig.~\ref{fig:pipeline}.
In this section, we start by introducing the unified  formulation.
\vspace{-.5em}
\subsection{Unified Formulation} \label{sec:definition}
\vspace{-.5em}
Given a video $V$ and a language query $Q$, we first divide $V$ into a sequence of $L_v$ fixed-length {\unit s} $\{v_1, \cdots, v_{L_v}\}$, where each \unit~$v_i$ is of length $l$ and has a centered timestamp $t_i$. The free-form text query $Q$ has $L_q$ tokens, denoted as $Q=\{q_1, \cdots, q_{L_q}\}$.
We then define three elements for each \unit~$v_i=\left( f_i, d_i, s_i \right)$, described as follows:
\begin{itemize}
    \vspace{-.5em}
    \item \textbf{Foreground indicator} $f_{i}\in\{0, 1\}$: a binary value indicating whether the $i$-th \unit~$v_i$ belongs to the foreground or not.
    If \unit~$v_i$ is the foreground of $Q$, then $f_i=1$, otherwise $f_i=0$.
    \vspace{-.5em}
    \item \textbf{Boundary offsets} $d_i=\left[d_i^s, d_i^e\right]\in \mathbb{R}^2$: 
    the temporal distance that converts the \unit~timestamp $t_i$ to its interval boundaries. Here, $d_i$ is valid when $f_i=1$. The $d_i^s$ is the distance between the starting of the interval and $t_i$, whereas $d_i^e$ is the distance between the ending and $t_i$.
    Thus, the whole temporal interval $b_i$ of $v_i$ can be represented as $b_i=[t_i-d_i^s, t_i+d_i^e]$
    \item \textbf{Saliency score} $s_i\in [0, 1]$: a continuous score determining the relevance between the visual content of \unit~$v_i$ and the query $Q$. 
    If the \unit~and query are highly correlated, $s_i=1$; 
    If they are totally irrelevant, then $s_i=0$. 
    Notably, it is reasonable to assume that $s_i>0$ if a clip is in the foreground of $Q$, otherwise $s_i=0$.
\end{itemize}

In Fig.\ref{fig:pipeline}~(a), we draw a schematic diagram to represent these three elements of clip $v_i$ in our definition.

\vspace{-.5em}
\subsection{Revisiting Various \vtg~Tasks and Labels}\label{sec:corpus}
\vspace{-.5em}
Treating \unit s as the atom composition of a video, we define the {\vtg~problem as collecting a target \unit~set $M=\{v_i\in V|Q\}$ from $V$, conditional on language query $Q$}.
We next illustrate how to extend this definition to various tasks and \data s. Especially, for each label, we answer:
\begin{enumerate}[itemsep=-1.pt, topsep=0pt] 
    \item How to collect scalable label corpus for pretraining?
    \item When using the unified formulation, how can we obtain unknown elements based on the available one?
\end{enumerate}

\vspace{-1.em}
\subsubsection{\MR~and \Interval~Label.}
\vspace{-.5em}

Moment retrieval aims to localize one~\cite{gao2017tall} or many~\cite{lei2020tvr} intervals in a video corresponding to \textit{a sentence} $Q$. 
As shown in Fig.~\ref{fig:pipeline} (Right blue), moment retrieval aims to select $m$ consecutive \unit~sets $M=M_1\cup \dots \cup M_m$, where $m\geq 1$, and 
$M_j$
is the $j$-th target moment.
$M$ can be simplified as the boundary set of foreground clips $\{b_i|f_i=1\}$.

The temporal \interval~with specific target boundaries is a common \data~for \mr.
However, annotating intervals requires manually reviewing the full video, which is expensive.
A solution is ASR~\cite{miech2019howto100m, yang2023vid2seq} that provide start and end timestamps, but ASR is often too noisy and poorly aligned with the visual content, making it suboptimal.
Here, we sought an alternative solution. 
We found that visual captions~\cite{sharma2018conceptual,bain2021frozen} tend to be descriptive, making them well-suited as grounding queries, thus if we can know how these videos are cut from the raw source, we can use this information to create pseudo intervals.
We find that VideoCC~\cite{nagrani2022learning} is a viable option for this purpose. It is worth noting that VideoCC is initially developed for video-level pretraining (\textit{e.g.,} power video-text retrieval), and we are the pioneer to investigate its potential in temporal grounding pretraining.

Once we obtain intervals, we convert \interval~\data s into the proposed formulation by defining $f_i=0$ and $s_i=0$ for \unit s that are not in target interval, and we assign $f_i=1$ and assume $s_i>0$ for \unit s that belongs to the target interval. 

\vspace{-.5em}
\noindent
\subsubsection{\HL~and \Curve~Label.}
\vspace{-.5em}
Highlight detection aims to assign an importance score to each video clip (making its annotations like a curve), then return the few highest-scoring clips as the highlight, where queries may~\cite{lei2020tvr} or may not~\cite{sun2014ranking, song2015tvsum} be provided as input.
For video highlighting datasets without language queries, we can use  \textit{video titles}~\cite{song2015tvsum} or \textit{video domain name}~\cite{sun2014ranking} as $Q$ because they are highly related to the topic of the video.
Then, this task is equivalent to picking \unit s with the top highest saliency scores \textit{i.e.} $M=\{v_i|s_i\in \text{top-}K\}$.

\begin{figure}[t]
    \includegraphics[width=1.0\linewidth]{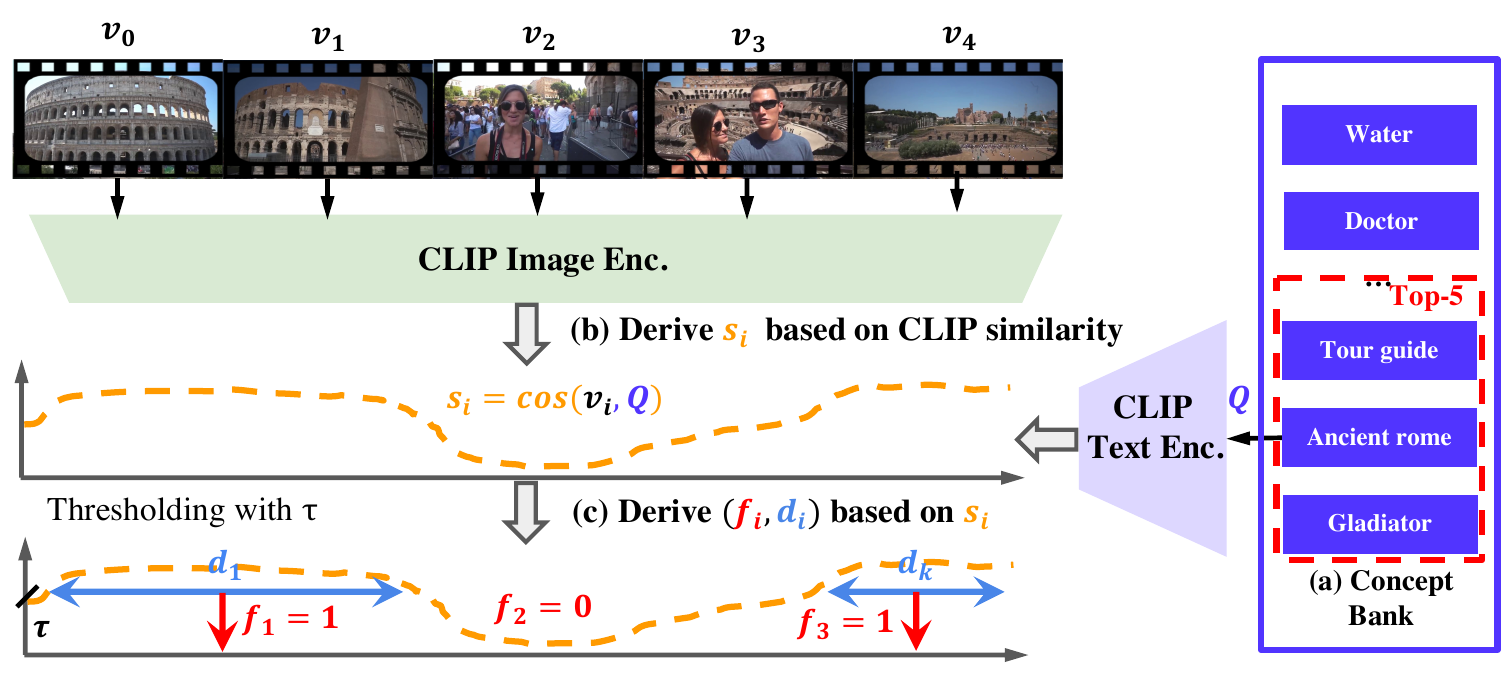}
    \captionsetup{font={small}}
    \caption{\small{
    \textbf{Process of using CLIP to produce temporal labels.}
    \textbf{(a)} We first use a concept bank to cover diverse open-world concepts. 
    \textbf{(b)} Next, we use CLIP as teacher to calculate the clip-level scores between each concept to get top-$5$ concepts as video gist, and treat their clip scores as saliency $s_i$. 
    \textbf{(c)} Based on $s_i$, we further derive the interval and point labels via thresholding.
    \vspace{-1em}
    }}
    \vspace{-1.em}
    \label{fig:demo4}
\end{figure}

Due to the interestingness contain subjectivity, the same video usually needs to be labeled by several people to eliminate bias. This makes \curve~labels the most expensive yet informative temporal annotations. 
Therefore, we are motivated to find an efficient way of producing scalable curve labels. 
Intuitively, interestingness reflects how each clip is relevant to the video gist. 
As depicted in Fig.~\ref{fig:demo4} (a),  we first define a concept bank using an open-world detection class list~\cite{shao2019objects365}.
Next, we use CLIP as a teacher to get the clip-level cosine similarities between each concept. Then, we select top-$5$ concepts as the video gist, and save their CLIP similarities as pseudo curve labels, \textit{i.e.,} Fig.~\ref{fig:demo4} (b).

As shown in Fig.~\ref{fig:demo4} (c), after obtaining curve labels, we assign $f_i=1$ for \unit s with $s_i$ greater than a threshold $\tau$, otherwise $f_i=0$. 
The $\tau$ is estimated based on the similarity of each video, refer to Supp. for details.
The offsets $d_i$ are defined as the distance between the foreground clip and its nearest neighboring clips where $f_i=0$.

\vspace{-1em}
\noindent
\subsubsection{\VS~and \Point~Label.}
\vspace{-.5em}
Query-focused \vsum~\cite{sharghi2017query} aims to summarize the entire video with a set of shots to provide a quick overview, with user-specific concepts (for example, trees and cars).
The generated summary should be succinct while representative of the entire video around the given query.
We define this task by regarding \textit{keywords} as $Q$, and select a set of \unit s $M=\{v_i|f_i=1\}$, where the size of $M$ is required to not exceed $\alpha\%$ of the original video length $|M| \leq \alpha\% |V|$ \textit{e.g.,} $\alpha=2\%$.

The annotations in QFVS~\cite{sharghi2017query} are point labels that indicate whether each shot belongs to the concept or not. The cost of  point labels is much cheaper than that of interval and curve labels since people only need to glance at a specific time. The recently \ego~\cite{grauman2022ego4d} dataset uses this point labeling to annotate massive-scale data by assigning a narration to an exact timestamp, such as ``I am opening the washing-machine'' at ${t}_i=2.30$ sec. Due to the favorable scale, it is natural to adapt them for large-scale pretraining. 
Recently, there have been attempts to improve video-text representation using point-wise annotations to improve the video-text representation~\cite{lin2022egocentric, zhao2023lavila, pramanick2023egovlpv2} and augment NLQ~\cite{grauman2022ego4d} baselines~\cite{ramakrishnan2023naq}.
Despite this, these methods mainly focus on transferring within the same domain.

For point labels, we derive $s_i>0$ if clip $f_i=1$, otherwise $s_i=0$. During pretraining, we estimate its temporal label $b_i$ based on the average distance between consecutive narrations within the video~\cite{lin2022egocentric, ramakrishnan2023naq, pramanick2023egovlpv2}.
\vspace{-.5em}
\section{Towards Unified VTG: Model} \label{sec:model}
\vspace{-.5em}
\noindent
We here introduce our unified model which seamlessly inherits our proposed unified formulation.

\vspace{-.5em}
\subsection{Overview}
\vspace{-.5em}
As shown in Fig.~\ref{fig:framework}, our model mainly comprises a frozen video encoder, a frozen text encoder, and a multi-modal encoder. The video and text encoders are keep consistent with Moment-DETR~\cite{lei2021detecting}, which employs the concatenation of CLIP~\cite{radford2021learning} (ViT-B/32) and SlowFast~\cite{feichtenhofer2019slowfast} (R-50) features as video representation, and use the CLIP text encoder~\cite{radford2021learning} to extract token level features.
Our multi-modal encoder  contains $k$ self-attention blocks that followed by three specific heads to decode the prediction.

Given an input video $V$ with $L_v$ clips and a language query $Q$ with $L_q$ tokens, we first apply the video encoder and the text encoder to encode the video and text respectively, then project them to the same dimension $D$ by two Feed-Forward Networks~(FFN), and thus obtain video features $\mathbf{V}=\{\mathbf{v}_i\}_{i=1}^{L_v}\in \mathbb{R}^{L_v\times D}$ and text features $\mathbf{Q}=\{\mathbf{q}_j\}_{j=1}^{L_q}\in \mathbb{R}^{L_q\times D}$.
Next, we design two pathways for cross-modal alignment and cross-modal interaction.

\textbf{(i)} For cross-modal alignment, 
we first adopt an attentive pooling operator to aggregate the query tokens $\mathbf{Q}\in \mathbb{R}^{L_q\times D}$ into a sentence representation $\mathbf{S}\in \mathbb{R}^{1\times D}$. Especially,
\begin{equation}
    \mathbf{S}=\mathbf{A}\mathbf{Q},
\end{equation}
where the weight $\mathbf{A}=\text{Softmax}\left(\mathbf{W}\mathbf{Q} \right)\in \mathbb{R}^{1\times L_q}$ and $\mathbf{W}^{1\times L_q}$ is a learnable embedding.
Then $\mathbf{V}$ and $\mathbf{S}$ are sent to perform contrastive learning (described in $\S$~\ref{sec:training_objective}).

\textbf{(ii)} For cross-modal interaction, learnable position embeddings $\mathbf{E}^{pos}$ and modality-type embeddings $\mathbf{E}^{type}$ are added to each modality to retain both positional and modality information:
\begin{equation}
\begin{aligned}
\tilde{\mathbf{V}} &= \mathbf{V} + \mathbf{E}^{pos}_V +  \mathbf{E}^{type}_V,\\
\tilde{\mathbf{Q}} &= \mathbf{Q} +  \mathbf{E}^{pos}_T+  \mathbf{E}^{type}_T.\\
\end{aligned}
\end{equation}

Next, the text and video tokens are concatenated and get a joint input $\mathbf{Z}^0=[\tilde{\mathbf{V}}; \tilde{\mathbf{Q}}]\in \mathbb{R}^{L\times D}$, where $L=L_v+L_q$. 
Further, $\mathbf{Z}^0$ is fed into the multi-modal encoder, which contains $k$ transformer layers with each layer consisting of a Multi-headed Self-Attention and FFN blocks.
\begin{equation}
\mathbf{Z}^d=\text{MLP}\left( \text{MSA} \left( \mathbf{Z}^{d-1} \right) \right),\quad d\in \{1\dots k\}.
\end{equation}
  
We take the video tokens $\tilde{\mathbf{V}}^k\in \mathbb{R}^{L_v\times D}$ from the multimodal encoder $E_{m}$ as output $\mathbf{Z}^k=[\tilde{\mathbf{V}^k}; \tilde{\mathbf{Q}^k}]\in \mathbb{R}$, and feed $\mathbf{Z}^k$ into the following heads for prediction.

\begin{figure}[t]
    \centering
    \vspace{-1.em}
    \includegraphics[width=0.65\linewidth]{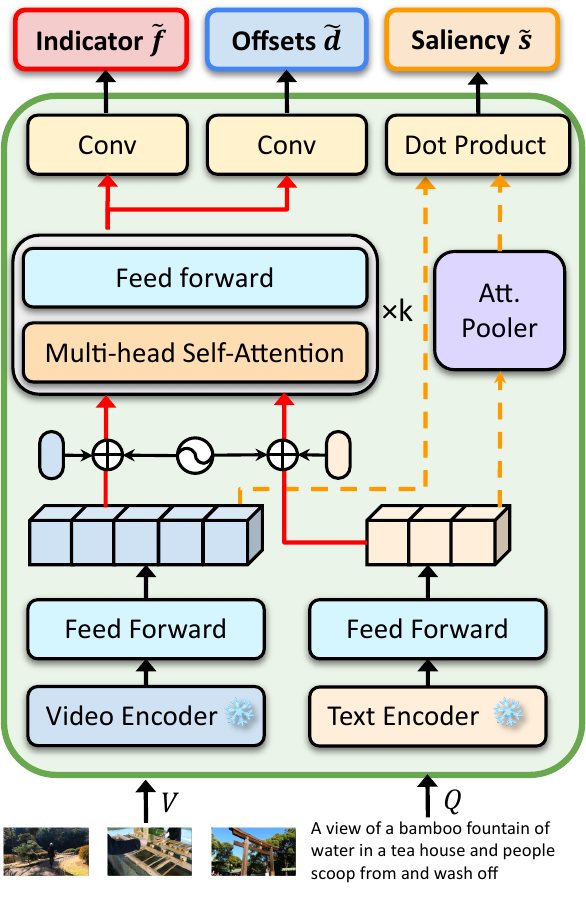}
    \captionsetup{font={small}}
    \caption{\small{\textbf{Unified grounding model} contains a video encoder, a text encoder, and a multi-modal encoder followed by three output heads, corresponding to three key elements $\left( \tilde{f}_i, \tilde{d}_i, \tilde{s}_i \right)$.
    Besides, our model has two pathways: one for cross-modal interaction (solid \textcolor{red}{red} line) and the other for cross-modal alignment (broken \textcolor{orange}{orange} line).
    \vspace{-1.0em}
    }}
    \label{fig:framework}
\end{figure}

\vspace{-.5em}
\subsection{Pretraining Objectives} \label{sec:training_objective}
\vspace{-.5em}
To match the previous unified formulation \textit{i.e.,} $\left( {f}_i, {d}_i, {s}_i \right)$, we devise three different heads to decode each element respectively, each one calling a capability.

\noindent
\textbf{Foreground head for Matching.}
Taking the output $\tilde{\mathbf{V}}^k\in \mathbb{R}^{L_v\times D}$ from the multi-modal encoder, this head applies three $1\times 3$ Conv layers, each with $D$ filters and followed by a ReLU activation. 
Finally, sigmoid activations are attached to output the prediction $\tilde{f}_i$ per \unit. 
We use the binary cross-entropy loss as a training objective.
\begin{equation}
\mathcal{L}_\text{f}=-\lambda_\text{f}\left( f_i \log \tilde{f}_i + \left(1-f_i\right)\log \left(1-\tilde{f_i}\right) \right).
\label{bce}
\end{equation}

\noindent
\textbf{Boundary head for Localization.} 
The design of this head is similar to the foreground head except for the last layer, which has $2$ outputs channel for the left and right offsets.
Taking the $\tilde{\mathbf{V}}^k\in \mathbb{R}^{L_v\times D}$, this head outputs offsets $\{\tilde{d}_i \}_i^{L_v}$ per \unit.
Then, we devise the predicted boundary $\tilde{b_i}$ and use the combination of smooth $L1$ loss~\cite{girshick2015fast} and generalized IoU loss~\cite{rezatofighi2019generalized} as our training objectives.
\begin{equation}
    \mathcal{L}_\text{b} = \mathbbm{1}_{f_i=1}  \left[
    \lambda_\text{L1}\mathcal{L}_{\text{SmoothL1}}\left(\tilde{d}_i, {d_i}\right)+
    \lambda_\text{iou} \mathcal{L}_\text{iou}\left( \tilde{b}_i, {b_i} \right) \right].
\end{equation}

Notably, this regression objective is only  devised for foreground \unit s~\textit{i.e.,} $f_i=1$. 

\noindent
\textbf{Saliency head for Contrasting.}
Since we define saliency as the relevance between visual context and text query, it is natural to interpret this score as a similar measurement between video and text modalities.
Taking the video tokens ${\mathbf{V}}=\{\mathbf{v}_i\}_{i=1}^{L_v}\in \mathbb{R}^{L_v\times D}$ and sentence representation $\mathbf{S}\in \mathbb{R}^{1\times D}$,
we define the predicted saliency score $\tilde{s}_i$ between \unit~$v_i$ and text query $Q$ as their cosine similarities:
\begin{equation}
\tilde{s}_i = \cos(\mathbf{v}_i, \mathbf{S}) := \frac{\mathbf{v}_i^T\mathbf{S}}{\|\mathbf{v}_i\|_2 \|\mathbf{S}\|_2},
\end{equation}
where $\|\cdot\|_2$ represents the $L2$-norm of a vector.

For each video $\mathbf{V}$, we randomly sample a foreground \unit~$\mathbf{v}_p$ with $f_p=1$ and $s_p>0$ as a positive sample; we treat other clips in the same video $\mathbf{v}_j$ with saliency $s_j$ less than $s_p$ as negative samples, \textit{i.e.,} $\Omega=\{j|s_j<s_p, 1 \leq j \leq {L}_v\}$, and perform \textbf{intra-video} contrastive learning:
\begin{equation}
\small{
\mathcal{L}_\text{s}^\text{intra}=-\log  \frac{\exp \left(\tilde{s}_p/\tau  \right)}{\exp \left(\tilde{s}_p/\tau  \right) + \sum_{j\in \Omega}\exp\left( \tilde{s}_j/ \tau \right)},
\label{intra}
}
\end{equation}
where $\tau$ is a temperature parameter and set as $0.07$.

Besides, 
we regard sentences from other samples within batches $k\in B$ as negative samples, and develop the \textbf{inter-video} contrastive learning for cross-sample supervision:
\begin{equation}
\mathcal{L}_\text{s}^\text{inter}=-\log\frac{\exp \left( \tilde{s}_p/\tau  \right)}{\sum_{k\in B}  \exp \left(\tilde{s}_p^k / \tau \right)},
\label{inter}
\end{equation}
where $B$ is the training batch size and $\tilde{s}_p^k=\cos (\mathbf{v}_i, \mathbf{S}_k)$.

Our saliency score head training loss is the combination of inter- and intra-video contrastive learning:
\begin{equation}
    \mathcal{L}_\text{s} = \lambda_\text{inter}\mathcal{L}_\text{s}^\text{inter} + \lambda_\text{intra}\mathcal{L}_\text{s}^\text{intra}.
\end{equation}

To this end, our total training objective is the combination of each head loss overall clips in the training set.
\begin{equation}
    \mathcal{L}=\frac{1}{N}\sum_{i=1}^N\left( \mathcal{L}_\text{f}+\mathcal{L}_\text{b}+\mathcal{L}_\text{s} \right),
\end{equation}
where $N$ is the clip number of the training set.

\subsection{Inference}
During inference, given a video $V$ and a language query $Q$, we first feed forward the model to obtain $\{ \tilde{f}_i, \tilde{b}_i, \tilde{s}_i\}_{i=1}^{L_v}$ for each \unit~$v_i$ from three heads.
Next, we describe how we carry out output for individual \vtg~tasks respectively.

\noindent
\textbf{\MR.}
We rank \unit s predicted boundaries $\{\tilde{b}_i\}_{i=1}^{L_v}$ based on their $\{\tilde{f}_i\}_{i=1}^{L_v}$ probabilities.
Since the predicted $L_v$ boundaries are dense, we adopt a 1-d Non-Max Suppression~(NMS) with a threshold $0.7$ to remove highly overlapping boundary boxes, yielding a final prediction.

\noindent
\textbf{\HL.}
For each \unit, to fully utilize the foreground and saliency terms, we rank all \unit s based on their $\{\tilde{f}_i+\tilde{s}_i\}_{i=1}^{L_v}$ scores, and then return the few top \unit~(\textit{e.g.,} Top-$1$) as predictions. 

\noindent
\textbf{\VS.}
Using the same preprocessing settings~\cite{sharghi2017query, xiao2020convolutional}, the videos are first divided as multiple segments via KTS algorithm~\cite{potapov2014category}. Then the clip scores from each segment are computed, and these scores are integrated. 
We rank all clips based on their foreground $\{\tilde{f}_i\}_{i=1}^{L_v}$ and return the Top-$2$\% clips as a video summary.

\begin{table}[t]
\centering
\scriptsize
\setlength{\tabcolsep}{1.5pt}
\vspace{-1.5em}
\begin{tabular}{lccccc}
	\toprule
    \textbf{Dataset}  & \textbf{Task}& \textbf{Pseudo?} & \textbf{Label} &  
    \textbf{\# Samples} &\textbf{Domain}\\
    \midrule
        Ego4D~\cite{grauman2022ego4d} & PT & \xmark & Point & $1.8$M & Egocentric \\
        VideoCC~\cite{nagrani2022learning} & PT & \cmark & Interval & $0.9$M & Web \\
        CLIP teacher~ & PT & \cmark & Curve & $1.5$M & Open \\
    \midrule
    QVHighlights~\cite{lei2021detecting} &  MR~+~HL & \xmark & Interval~+~Curve & $10.3$K & VLog, News \\
    NLQ~\cite{grauman2022ego4d} & MR  & \xmark &Interval & $15.1$K & Egocentric \\
    Charades-STA~\cite{gao2017tall} & MR  & \xmark &Interval & $16.1$K & Indoor \\
    TACoS~\cite{regneri2013grounding} & MR & \xmark & Interval & $18.2$K & Kitchens \\
    YoutubeHL~\cite{sun2014ranking} & HL & \xmark & Curve & $600$  & Web \\
    TVSum~\cite{song2015tvsum} & HL & \xmark & Curve & $50$ & Web \\
   QFVS~\cite{sharghi2017query} & VS & \xmark & Point & $4$ & Egocentric \\
\bottomrule
\end{tabular}
\centering
\captionsetup{font={small}}
\caption{
\small{
\textbf{Dataset statistics. }
\textbf{The upper side} datasets are used for pretraining (PT) which cover three label types, two of which are pseudo. 
\textbf{The lower side} datasets are used for downstream tasks (MR:~\MR, HL:~\HL, VS:~\VS).
}}
\vspace{-1em}
\label{tab:dset}
\end{table}
\vspace{-1em}
\section{Experiments}
\vspace{-.5em}
In this section, we conduct experiments on various benchmarks to evaluate our approach. 
Mainly, we design the experiments to study the following questions:

\noindent{$\mathbf{Q1}$:} How much improvement could be made by \our~grounding pretraining?

\noindent{$\mathbf{Q2}$:} What are the effects of using different pretraining corpus from various labels?

\noindent{$\mathbf{Q3}$:} Is it necessary to use the proposed unified formulation and unified model?

More ablation studies can be found in \supp.

\vspace{-.5em}
\subsection{Datasets and Settings}
\vspace{-.5em}
We have summarized the dataset information in Tab.\ref{tab:dset}.
For pretraining, we gather $1.8$M \point~labels from \ego~and $0.9$M \interval~labels from VideoCC~\cite{nagrani2022learning}. For curve labels, we apply CLIP teacher method (Fig.~\ref{fig:demo4}) to \ego~and VideoCC datasets to get $1.5$M pseudo labels. Therefore, a total of $4.2$M temporal annotations are used for grounding pretraining.
For downstream tasks, we assess our methods on four \vtg~tasks across seven datasets, spanning 
\textbf{(i)} Jointly \mr~and \hl;
\textbf{(ii)} \MR;
\textbf{(iii)} \HL;
\textbf{(iv)} \VS.
Additional details are listed in Supp.

\noindent
\textbf{Evaluation Metrics.}
For \qv, we follow official \cite{lei2020tvr}, Recall@$1$ with IoU thresholds $0.5$ and $0.7$, mean average precision (mAP) with IoU thresholds $0.5$ and $0.75$, and the average mAP over a series of IoU thresholds $[0.5$:$0.05$:$0.95]$ are used for \mr. For \hl, mAP and HIT@$1$ are used, a clip is treated as a true positive if it has the saliency score of {Very Good}. 
For \charades, NLQ, \tacos, Recall@$1$ with IoU thresholds $0.3$, $0.5$ and $0.7$, and mIoU are used. 
For \youtube~and \tvsum, we follow \cite{liu2022umt} and use mAP and Top-$5$ mAP, respectively.
For \qfvs, we follow \cite{wu2022intentvizor} that reports F1-score per video as well as an average.

\noindent
\textbf{Implementation Details.}
We set $k=4$ multi-modal transformer encoder layers, with $d=1024$ hidden size and $8$ attention heads. 
The drop path rates are $0.1$ for transformer layers and $0.5$ for input FFN projectors. 
During the pretraining stage, our experiments are carried out on $8$ A100 GPUs. 
When it comes to downstream tasks, we use one GPU.
For \mr, all baselines and \our~use the same video and text features. For \hl~and \vsum, we report results following~\cite{liu2022umt} and~\cite{wu2022intentvizor}.
See Supp. for more details.

\begin{table}[t]
\footnotesize
\setlength{\tabcolsep}{0pt}
\begin{tabularx}{\linewidth}{@{\hspace{0mm}}p{2.2cm}p{0.85cm}<{\centering}p{0.85cm}<{\centering}p{1.0mm}<{\centering}p{0.85cm}<{\centering}p{0.85cm}<{\centering}p{0.85cm}<{\centering}p{1.0mm}<{\centering}p{0.85cm}<{\centering}p{0.8cm}<{\centering}}
\toprule
& \multicolumn{6}{c}{\textbf{\MR}} & & \multicolumn{2}{c}{\textbf{HD}} \\
\cmidrule{2-7} \cmidrule{9-10}
& \multicolumn{2}{c}{R$1$} & & \multicolumn{3}{c}{mAP} & & \multicolumn{2}{c}{$\geq$ Very Good} \\
\cmidrule{2-3} \cmidrule{5-7} \cmidrule{9-10}
\vspace{-0.73cm}\hspace{0.8cm}\textbf{Method} & @$0.5$ & @$0.7$ & & @$0.5$ & @$0.75$ & Avg. & & mAP & HIT@$1$ \\
\midrule
BeautyThumb \cite{song2016click} & $-$  & $-$  & & $-$  & $-$ & $-$  & & $14.36$ & $20.88$ \\
DVSE \cite{liu2015multi} & $-$ & $-$ & & $-$ & $-$ & $-$ & & $18.75$ & $21.79$ \\
MCN \cite{anne2017localizing} & $11.41$ & $2.72$ & & $24.94$ & $8.22$ & $10.67$ & & $-$ & $-$ \\
CAL \cite{escorcia2019temporal} & $25.49$ & $11.54$ & & $23.40$ & $7.65$ & $9.89$ & & $-$  & $-$  \\
CLIP \cite{radford2021learning} & $16.88$ & $5.19$ & & $18.11$ & $7.0$ & $7.67$ & & $31.30$  & $61.04$  \\
XML \cite{lei2020tvr} & $41.83$ & $30.35$ & & $44.63$ & $31.73$ & $32.14$ & & $34.49$ & $55.25$ \\
XML+ \cite{lei2021detecting} & $46.69$ & $33.46$ & & $47.89$ & $34.67$ & $34.90$ & & $35.38$ & $55.06$ \\
\midrule
MDETR \cite{lei2021detecting} & $52.89$ & $33.02$ & & $54.82$ & $29.40$ & $30.73$ & & $35.69$ & $55.60$ \\
MDETR \ \ w/ PT & $59.78$ & $40.33$ & & ${60.51}$ & $35.36$ & $36.14$ & & $37.43$ & $60.17$ \\
{UMT}$\dagger$\cite{liu2022umt} & $56.23$ & $41.18$ & & $53.83$ & $37.01$ & $36.12$ & & $38.18$ & $59.99$ \\
{UMT}$\dagger$ \ \ \ \ \ \ w/ PT & ${60.83}$ & ${43.26}$ & & $57.33$ & ${39.12}$ & ${38.08}$ & & ${39.12}$ & ${62.39}$ \\
\midrule 
{\our} & $58.86$ & $40.86$ & & $57.60$ & $35.59$ & $35.47$ & & $38.20$ & $60.96$ \\
{\our} \ w/ PT & $\mathbf{65.43}$ & $\mathbf{50.06}$ & & $\mathbf{64.06}$ & $\mathbf{45.02}$ & $\mathbf{43.63}$ & & $\mathbf{40.54}$ & $\mathbf{66.28}$ \\
\rowcolor{Gray}
{\our} ZS & $25.16$ & $8.95$ & & $27.42$ & $7.64$ & $10.87$ & & $35.96$ & $53.50$ \\
\bottomrule
\end{tabularx}
\captionsetup{font={small}}
\caption{\textbf{Jointly \MR~and \HL~results on QVHighlights test split\protect\footnotemark.} $\dagger$: introduce audio modality. w/ PT: fine-tuning after pre-training; ZS: zero-shot inference.}
\vspace{-2em}
\label{tab:qvhl}
\end{table}
\footnotetext{\href{https://codalab.lisn.upsaclay.fr/competitions/6937\#results}{Codalab QVHighlights Evaluation}}
\begin{table*}[!b]
\vspace{-.5em}
\footnotesize
\centering
\begin{tabular}{l|cccc|cccc|cccc}
	\toprule
	\multicolumn{1}{c|}{\multirow{2}{*}{{\textbf{Method}}}}  &  \multicolumn{4}{c|}{{\textbf{NLQ}~\cite{grauman2022ego4d}}} & \multicolumn{4}{c|}{{\textbf{Charades-STA}~\cite{gao2017tall}}} & \multicolumn{4}{c}{{\textbf{TACoS}~\cite{regneri2013grounding}}} \\
	& R@$0.3$ & R@$0.5$ & R@$0.7$ & mIoU &    R@$0.3$ & R@$0.5$ & R@$0.7$ & mIoU &  R@$0.3$ & R@$0.5$ & R@$0.7$ & mIoU\\  \midrule[1pt] 
 2D TAN~\cite{zhang2020learning} & $4.33$ &  $1.83$ &  $0.60$ &  $3.39$ & $58.76$ & $46.02$ & $27.50$ & $41.25$ & $40.01$ & $27.99$ & $12.92$ & $27.22$\\
 VSLNet~\cite{zhang2020span} & $4.54$ & $2.40$ & $1.01$ & $3.54$ & $60.30$ & $42.69$ & $24.14$ & $41.58$ & $35.54$ & $23.54$ & $13.15$ & $24.99$ \\
MDETR~\cite{lei2021detecting} & $4.34$ & $1.81$ & $0.65$ & $3.53$ & $65.83$& $52.07$ & $30.59$ & $45.54$& $37.97$ & $24.67$ & $11.97$ & $25.49$\\
\midrule
{\our} &  $7.28$ & $3.95$ & $1.32$ & $4.91$ & $70.81$ & $58.01$ &  $35.65$ & $50.10$ & $51.44$ & $34.97$ & $17.35$ & $33.60$ \\ 
{\our} w/ PT & $\mathbf{11.74}$ & $\mathbf{7.54}$ & $\mathbf{3.25}$ & $\mathbf{7.88}$ & $\mathbf{72.63}$ & $\mathbf{60.19}$ &  $\mathbf{38.55}$ & $\mathbf{52.17}$ & $\mathbf{56.11}$ & $\mathbf{43.44}$ & $\mathbf{24.27}$ & $\mathbf{38.63}$ \\ 
\rowcolor{Gray}
{\our} ZS & $6.48$ & $3.48$ & $1.16$ & $4.63$ & $44.09$ & $25.22$ &  $10.03$ & $27.12$ & $5.17$ & $1.27$ & $0.27$ & $4.40$ \\ 
\bottomrule
\end{tabular}
\centering
\vspace{-1em}
\captionsetup{font={small}}
\caption{\small{\textbf{\MR~results on NLQ, Charades-STA, and TACoS benchmarks.}
All baselines use the same video features~(CLIP ViT-B/32 and SlowFast R-50) and text features~(CLIP text enc.). 
w/ PT means fine-tuning after pre-training;
ZS means zero-shot inference.  
\vspace{-1em}
}}
\label{tab:mr}
\end{table*}

\subsection{Comparison with State-of-the-arts ($\mathbf{Q1}$)}
\subsubsection{Joint \MR~and \HL}
\vspace{-.5em}
As illustrated in Tab.~\ref{tab:qvhl}, we first evaluate our \our~on {\qv}~{test} split: 
{\text{(i)}} Without pretraining, \our~has shown comparable performance to two joint optimization counterparts \detr~\cite{lei2021detecting} and UMT~\cite{liu2022umt}, demonstrating its superior model design for joint task optimization. 
{\text{(ii)}} With large-scale pretraining, \our~exhibits a significant improvement on all metrics, such as {${+8.16}$ Avg. mAP and ${+5.32}$ HIT@$1$}. As a result, \our~surpasses all baselines by a large margin. Notably, UMT introduces audio modality and ASR pretraining~\cite{liu2022umt}, but it is still worse than us by Avg. mAP of ${5.55}$ and HIT@$1$ of ${3.89}$.
{\text{(iii)}} Due to the large-scale pretraining, \our~can perform {zero-shot} grounding and outperforms several supervised baselines without any training samples.

\vspace{-1em}
\subsubsection{\MR}
\vspace{-.5em}
In Tab.~\ref{tab:mr}, we compare the results of our method and the mainstream \mr~methods on three widely used benchmarks. 
\text{(i)} Similar to the observation made by \qv, without pretraining,
we find that \our~is still superior to other compared methods. 
This demonstrates once more the effectiveness of our concise architecture.
\text{(ii)} Large-scale grounding pretraining has resulted in significant improvements, leading to a considerable increase in the mIoU \textit{i.e.,} $+2.97$ in NLQ, $+2.07$ in \charades, and $+5.03$ in \tacos. 
\text{(iii)}
Notably, in NLQ, our zero-shot result has outperformed all the baselines methods due to the close pretraining domain.
However, it is worth mentioning that the zero-shot performance on \tacos~is inferior. This could be because the videos have scenes that are very similar to each other, with only small spatial variations, making it difficult to effectively apply zero-shot methods.

\vspace{-1em}
\subsubsection{\HL}
\vspace{-.5em}
In Tab.~\ref{tab:youtube} and Tab.~\ref{tab:tvsum}, we conduct \hl~experiments on \youtube~and \tvsum~respectively,  where the baselines with $\dagger$ (rows 6-9) are incorporate with audio features.
We observe that {(i)} grounding pretraining brings  improvement on \our~and surpasses all baselines in Avg. mAP. 
(ii) In \tvsum, gain discrepancy among domains may stem from its small scale (50 samples) and scoring subjectivity. In contrast, the larger YouTube dataset (600 videos) yields more consistent pretraining gains.
{(ii)} Moreover, in {zero-shot} setting, \our~beats several video-only baselines such as \cite{sun2014ranking, wang2020trailer}.
\begin{figure*}[t]
 \begin{minipage}[t]{0.4\textwidth}
  \centering
  \scriptsize
\renewcommand\tabcolsep{0pt}
\footnotesize
\begin{tabularx}{\linewidth}{@{\hspace{1mm}}p{2cm}|@{\hspace{0.5mm}}p{0.7cm}<{\centering}p{0.7cm}<{\centering}p{0.7cm}<{\centering}p{0.7cm}<{\centering}p{0.7cm}<{\centering}p{0.7cm}<{\centering}p{0.7cm}<{\centering}}
\toprule
\textbf{Method} & {Dog} & {Gym.} & {Par.} & {Ska.} & {Ski.} & {Sur.} & \textbf{Avg.} \\
\midrule
RRAE \cite{yang2015unsupervised} & $49.0$ & $35.0$ & $50.0$ & $25.0$ & $22.0$ & $49.0$ & $38.3$ \\
GIFs \cite{gygli2016video2gif} & $30.8$ & $33.5$ & $54.0$ & $55.4$ & $32.8$ & $54.1$ & $46.4$ \\
LSVM \cite{sun2014ranking} & $60.0$ & $41.0$ & $61.0$ & $62.0$ & $36.0$ & $61.0$ & $53.6$ \\
LIM-S \cite{xiong2019less} & $57.9$ & $41.7$ & $67.0$ & $57.8$ & $48.6$ & $65.1$ & $56.4$ \\
SL-Module \cite{xu2021cross} & ${70.8}$ & ${53.2}$ & ${77.2}$ & ${72.5}$ & ${66.1}$ &${76.2}$ & ${69.3}$ \\\midrule
MINI-Net$\dagger$ \cite{hong2020mini} & $58.2$ & $61.7$ & $70.2$ & $72.2$ & $58.7$ & $65.1$ & $64.4$ \\
TCG$\dagger$ \cite{ye2021temporal} & $55.4$ & $62.7$ & $70.9$ & $69.1$ & $60.1$ & $59.8$ & $63.0$ \\
Joint-VA$\dagger$ \cite{badamdorj2021joint} & $64.5$ & $71.9$ & $80.8$ & $62.0$ & $73.2$ & $78.3$ & $71.8$ \\
{UMT}$\dagger$\cite{liu2022umt} & $65.9$ & $75.2$ & $\mathbf{81.6}$ & $71.8$ & $72.3$ & $82.7$ & $74.9$ \\ \midrule
{\our} &  $71.8$ & $76.5$ & $73.9$ & $73.3$ & $73.2$ & $82.2$ & $75.2$ \\ 
{\our} w/ PT &  $\mathbf{74.3}$ & $\mathbf{79.0}$ & $74.4$ & $\mathbf{84.9}$ & $\mathbf{75.1}$ & $\mathbf{83.9}$ & $\mathbf{78.6}$  \\
\rowcolor{Gray}
{\our} ZS &  $36.8$ & $62.8$ & $65.9$ & $39.2$ & $64.5$ & $54.0$ & $53.9$  \\
\bottomrule
\end{tabularx}
\vspace{-1em}
\captionsetup{font={small}}
\makeatletter\def\@captype{table}\makeatother\caption{\small{\textbf{\HL~results of mAP on YouTube HL.} $\dagger$ denotes using audio modality.}}
\label{tab:youtube}
  \end{minipage}
  \hspace{0.25cm}
  \begin{minipage}[t]{0.565\textwidth}
   \centering
   \scriptsize
\renewcommand\tabcolsep{0pt}
\footnotesize
\begin{tabularx}{\linewidth}{@{\hspace{1mm}}p{2cm}|@{\hspace{0.5mm}}p{0.9cm}<{\centering}p{0.7cm}<{\centering}p{0.7cm}<{\centering}p{0.7cm}<{\centering}p{0.7cm}<{\centering}p{0.7cm}<{\centering}p{0.7cm}<{\centering}p{0.7cm}<{\centering}p{0.7cm}<{\centering}p{0.7cm}<{\centering}p{0.7cm}<{\centering}}
\toprule
\textbf{Method} &  {VT} & {VU} & {GA} & {MS} & {PK} & {PR} & {FM} & {BK} & {BT} & {DS} & \textbf{Avg.} \\
\midrule
sLSTM \cite{zhang2016video} & ${41.1}$ & ${46.2}$ & $46.3$ & $47.7$ & $44.8$ & $46.1$ & $45.2$ & $40.6$ & $47.1$ & $45.5$ & $45.1$ \\
SG \cite{mahasseni2017unsupervised} & $42.3$ & $47.2$ & $47.5$ & $48.9$ & $45.6$ & $47.3$ & $46.4$ & $41.7$ & $48.3$ & $46.6$ & $46.2$ \\
LIM-S \cite{xiong2019less} & $55.9$ & $42.9$ & $61.2$ & $54.0$ & $60.4$ & $47.5$ & $43.2$ & $66.3$ & $69.1$ & $62.6$ & $56.3$ \\
Trailer \cite{wang2020trailer} &$61.3$ &$54.6$ & $65.7$ & $60.8$ & $59.1$ & ${70.1}$ & $58.2$ & $64.7$ & $65.6$ & ${68.1}$ & $62.8$ \\
SL-Module \cite{xu2021cross} & ${86.5}$ & ${68.7}$ & ${74.9}$ & $\mathbf{86.2}$ & ${79.0}$ & $63.2$ & ${58.9}$ & ${72.6}$ & ${78.9}$ & $64.0$ & ${73.3}$ \\\midrule
MINI-Net$\dagger$ \cite{hong2020mini} & $80.6$ & $68.3$ & $78.2$ & $81.8$ & $78.1$ & $65.8$ & $57.8$ & $75.0$ & $80.2$ & $65.5$ & $73.2$ \\
TCG$\dagger$ \cite{ye2021temporal} & $85.0$ & $71.4$ & $81.9$ & $78.6$ & $80.2$ & $75.5$ & $71.6$ & $77.3$ & $78.6$ & $68.1$ & $76.8$ \\
Joint-VA$\dagger$ \cite{badamdorj2021joint} & $83.7$ & $57.3$ & $78.5$ & $86.1$ & $80.1$ & $69.2$ & $70.0$ & $73.0$ & $\mathbf{97.4}$ & $67.5$ & $76.3$ \\
{UMT}$\dagger$\cite{liu2022umt} & ${87.5}$ & ${81.5}$ & ${88.2}$ & $78.8$ & ${81.5}$ & $\mathbf{87.0}$ & $\mathbf{76.0}$ & ${86.9}$ & ${84.4}$ & $\mathbf{79.6}$ &
${83.1}$ \\
\midrule
{\our} &  $83.9$ & $\mathbf{85.1}$ & $89.0$ & $80.1$ & $\mathbf{84.6}$ & $81.4$ & $70.9$ & $91.7$ & $73.5$ & $69.3$ & $81.0$ \\
{\our} w/ PT & $\mathbf{92.0}$ & ${77.8}$ & $\mathbf{89.8}$ & ${83.8}$ & ${82.2}$ & ${85.8}$ & ${74.3}$ & $\mathbf{91.8}$ & ${90.5}$ & ${77.6}$ & $\mathbf{84.6}$\\
\rowcolor{Gray}
{\our} ZS &  ${78.5}$ & ${67.0}$ & ${75.3}$ & ${63.6}$ & ${67.0}$ & ${66.8}$ & ${35.4}$ & ${85.3}$ & ${83.1}$ & ${50.0}$ & ${67.2}$\\
\bottomrule
\end{tabularx}
\vspace{-1em}
\captionsetup{font={small}}
\makeatletter\def\@captype{table}\makeatother\caption{\small{\textbf{\HL~results of Top-5 mAP on TVSum.} $\dagger$ denotes using audio modality.}}
\label{tab:tvsum}
\end{minipage}
\vspace{-1em}
\end{figure*}

\vspace{-3em}
\subsubsection{\VS}
\vspace{-.5em}
In Tab.~\ref{tab:qfvs}, we present the \qfvs~benchmark results. Our pretrained \our~achieves a $0.8\%$ higher Avg. F1-score than \iv~\cite{wu2022intentvizor}, where the latter is an interactive method and being tailored for the \vsum~task. 
This result demonstrates the generalization of our method on \vsum~task.
\begin{table}[h]
\footnotesize
\centering
\vspace{-.5em}
\begin{tabular}{l|ccccc}
	\toprule
	\multicolumn{1}{c|}{\multirow{1}{*}{\textbf{Method}}}  & V$1$ & V$2$ & V$3$ & V$4$ & \textbf{Avg.} \\ \midrule
 	QC-DPP~\cite{sharghi2017query} & $48.68$ & $41.66$ & $36.51$ & $29.96$ & $44.19$ \\
 	CHAN~\cite{xiao2020convolutional} & $49.14$ & $46.53$ & $58.65$ & $33.42$ & $46.94$ \\
 	QSAN~\cite{xiao2020query} & $48.52$ & $46.64$ & $56.93$ & $34.25$ & $46.59$ \\
  	WHM~\cite{nalla2020watch} & ${50.96}$ & $48.28$ & ${58.41}$ & ${39.18}$ & $49.20$ \\
 	IntentVizor~\cite{wu2022intentvizor} & ${51.27}$ & ${53.48}$ & $\mathbf{61.58}$ & ${37.25}$ & ${50.90}$ \\ \midrule
{\our} & $\mathbf{52.54}$ & ${54.48}$ & ${56.73}$ & ${40.37}$ & ${51.03}$ \\ 
{\our} w/ PT & ${49.85}$ & $\mathbf{56.97}$ & $59.35$ & $\mathbf{40.62}$ & $\mathbf{51.70}$ \\
\bottomrule
\end{tabular}
\centering
\vspace{-.5em}
\captionsetup{font={small}}
\caption{\small{\textbf{Video Summarization results of F-score on \qfvs.}
}}

\vspace{-1em}
\label{tab:qfvs}
\end{table}

\begin{table}[!b]
\centering
\scriptsize
\setlength{\tabcolsep}{1pt}
\begin{tabular}{c|ccc|ccc|cc|c|c}
\toprule
 & \multicolumn{3}{c|}{\textbf{Pretraining Corpus}} &
  \multicolumn{3}{c|}{\textbf{Unified Labels?}} &
  \multicolumn{2}{c|}{\textbf{QVHighlights}} &
  \multicolumn{1}{c|}{\textbf{TACoS}} & 
  \multicolumn{1}{c}{\textbf{YouTube}}\\
row& 
Ego4D &
  VideoCC &
  CLIP &
  \multirow{2}{*}{{Point}} &
  \multirow{2}{*}{{Interval}} &
  \multirow{2}{*}{{Curve}} &
  \multicolumn{1}{c}{MR} &
  \multicolumn{1}{c|}{HL} &
  \multicolumn{1}{c|}{MR} & 
  \multicolumn{1}{c}{HL} \\
  & {Point} & {Interval} & {Curve}  &  & & &  mAP & mAP &  mIoU & mAP \\ \midrule
${1}$ &  &  &   &  &  &  &   $36.13$  &  $38.83$  &  $33.60$ & $75.15$  \\ \midrule
${2}$ & \cmark &   &   & \cmark &  &  &  $39.89$ & $39.48$ &  $35.33$    & $75.32$ \\  
${3}$ &  & \cmark  &   &  & \cmark &  &  $39.81$  & $39.75$ &  $35.11$    &  $74.76$\\  
${4}$ &  &   &  \cmark &  & & \cmark &  $39.16$  & $39.80$  &   $35.68$   & $75.44$ \\ 
${5}$ & \cmark & \cmark  &  \cmark & \cmark & \cmark & \cmark & $41.37$ & $39.97$ & $35.87$ &  $75.66$  \\  
 \midrule
${6}$ & \cmark &   &   & {\cmark} & \underline{\cmark} & \underline{\cmark}  &   $41.53$  & $39.66$ &  $36.52$    & $75.27$ \\  
$7$ &  & \cmark  &   & \underline{\cmark} & \cmark & \underline{\cmark}     & $40.96$  & $40.10$ &  $36.78$    & $76.10$ \\  
${8}$ &  &   &  \cmark & \underline{\cmark} & \underline{\cmark} & \cmark    & $42.19$ &  $40.43$  & $35.85$  & $77.48$  \\  
$9$ & \cmark & \cmark  &  \cmark & \underline{\cmark} & \underline{\cmark} & \underline{\cmark} &  $\mathbf{45.99}$ & $\mathbf{41.25}$ & $\mathbf{38.63}$ & $\mathbf{78.56}$   \\  \midrule
\end{tabular}
\centering
\vspace{-1em}
\captionsetup{font={small}}
\caption{
\small{\textbf{Ablation studies on pretraining corpus.} \underline{\cmark} denotes the elements derived by us, which are not provided in vanilla training corpus: Ego4D, VideoCC, and CLIP teacher.}}
\label{tab:q2q3}
\end{table}

\subsection{Ablation Studies}
\noindent \textbf{Effect of different labels for pretraining ($\mathbf{Q2}$).}
In Tab.~\ref{tab:q2q3} top half, we investigate the effect of different labels corpus for pretraining. The results here are before unified formulation \textit{i.e.,} the original label provided by the pretraining set.
Our findings ({rows} 1-4) indicate that (i) incorporating any type of label for pretraining yields considerable performance gains on most benchmarks. 
(ii) Combining all three types of data ({row} 5) for pretraining further boost the outcomes, such as $+5.2$ MR's mAP and $+1.1$ HL's mAP over baseline ({row} 1) on \qv.
 
\noindent \textbf{Effect of unified formulation ($\mathbf{Q3}$).}
In Tab.~\ref{tab:q2q3} bottom half, we further study the impacts of unified formulation \textit{i.e.,} the benefits of deriving unknown elements for pretraining.
From {rows} 2-4 vs {rows} 6-8, We find that (i) training corpora receive performance gains in most settings, 
which proves that the label converting methods are crucial for better utilizing temporal labels.
(ii) Among all settings, curve labels appear to be the most effective ones, and beat the manual point labels except in a few domains \textit{e.g.,} TACoS.
(iii) We get the optimal result ({row} 9) by using full three converted corpus for pretraining, with $4.62$ MR's mAP and $1.28$ HL's mAP increase over counterparts ({row} 5) on QVHighlights.

\noindent \textbf{Effect or pretraining scale.}
In Fig.~\ref{fig:scale}, we explore the effect of utilizing various scales of labels for pretraining. We observe a steady performance improvement on both \mr~and \hl~tasks as the training sample size increases.  It also shows that unifying labels to construct a large training corpus can greatly benefit the \vtg.

\begin{figure}
  \centering
  \begin{minipage}[t]{0.49\linewidth}
  \centering
\begin{tikzpicture}
	\begin{axis} [
		axis x line*=bottom,
		axis y line*=left,
		legend pos=north east,
		ymin=40, ymax=56,
		xmin=0, xmax=100,
		xticklabel={\pgfmathparse{\tick}\pgfmathprintnumber{\pgfmathresult}\%},
		xtick={0,25,50,75,100},
		ytick={40, 44, 48, 52, 56},
         ylabel={R$1$@$0.7$},
            yticklabel pos=left,
            ylabel style={font=\Huge, yshift=-15pt},
		xlabel style={font=\Huge},
		width=\linewidth,
		legend style={cells={align=left}},
		label style={font=\footnotesize},
		tick label style={font=\footnotesize},
		legend style={at={(0.35,0.2)},anchor=west},
		]
		\addplot[mark=triangle,style={thick},NiceBlue] plot coordinates {
        (0, 40.9)
        (25, 46.39)
        (50, 47.68)
        (75, 48.77)
        (100, 54.00)
		};
	\end{axis}
    \end{tikzpicture}
\captionsetup{labelformat=empty}
\vspace*{-1em}
\caption*{\scriptsize{(a) R$1$@$0.7$ of \MR. \vspace{-1em}}}
  \end{minipage}
  \begin{minipage}[t]{0.49\linewidth}
  \centering
\begin{tikzpicture}
 	\begin{axis} [
		axis x line*=bottom,
		axis y line=left,
		legend pos=north east,
		ymin=38, ymax=42,
		xmin=0, xmax=100,
		xticklabel={\pgfmathparse{\tick}\pgfmathprintnumber{\pgfmathresult}\%},
		xtick={0,25,50,75,100},
		ytick={38, 39, 40, 41, 42},
            ylabel={mAP Avg.},
            yticklabel pos=right,
            ylabel style={font=\Huge, yshift=-15pt},
		xlabel style={font=\Huge},
		width=\linewidth,
		legend style={cells={align=left}},
		label style={font=\footnotesize},
		tick label style={font=\footnotesize},
		legend style={at={(0.35,0.2)},anchor=west},
		]

		\addplot[mark=o,style={thick},orange] plot coordinates {
        (0, 38.83)
        (25, 39.95)
        (50, 40.36)
        (75, 40.39)%
        (100, 41.25)
		};
	\end{axis}
\end{tikzpicture}
\captionsetup{labelformat=empty}
\vspace*{-1em}
\caption*{\scriptsize{(b) mAP Avg. of \HL.  \vspace{-1em}}}  
  \end{minipage}
  \captionsetup{font={small}}
  \caption{\small{\textbf{Effect of pretraining scale on \qv~dataset.}}}
  \vspace{-2em}
    \label{fig:scale}
\end{figure}

\vspace{-1.em}
\section{Conclusion}
\vspace{-.5em}
This paper introduces \our, a framework that unifies diverse \vtg~tasks and \data s by addressing three key challenges:
(i) We define a unified formulation for \vtg~to convert various \data s and tasks under a single framework, and propose a label scaling scheme.
(ii) We develop an effective yet flexible model that can handle various \vtg~tasks and training \data s.
(iii) Due to the unified framework and availability of scalable labels, it becomes feasible to perform large-scale temporal grounding pretraining over diverse labels.
We demonstrate the effectiveness and flexibility of our \our~on four settings across seven datasets, spanning joint optimization as well as individual tasks.

\clearpage
\begin{figure*}[h]
    \begin{minipage}{\textwidth}
        \flushleft
        \captionsetup{labelformat=empty, justification=raggedright,singlelinecheck=false}
        \caption*{\large{\textbf{(a) \qv:} \textit{Vlog and News} domains, videos are average 2.5 minutes long; Each video might have several intervals}}
        \includegraphics[width=\textwidth]{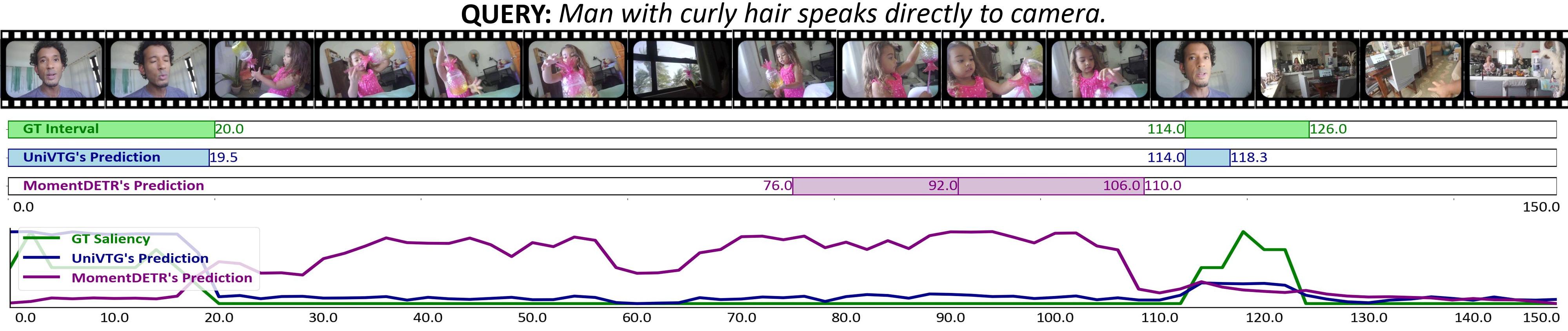}
    \end{minipage}
    \begin{minipage}{\textwidth}
        \centering
        \includegraphics[width=\textwidth]{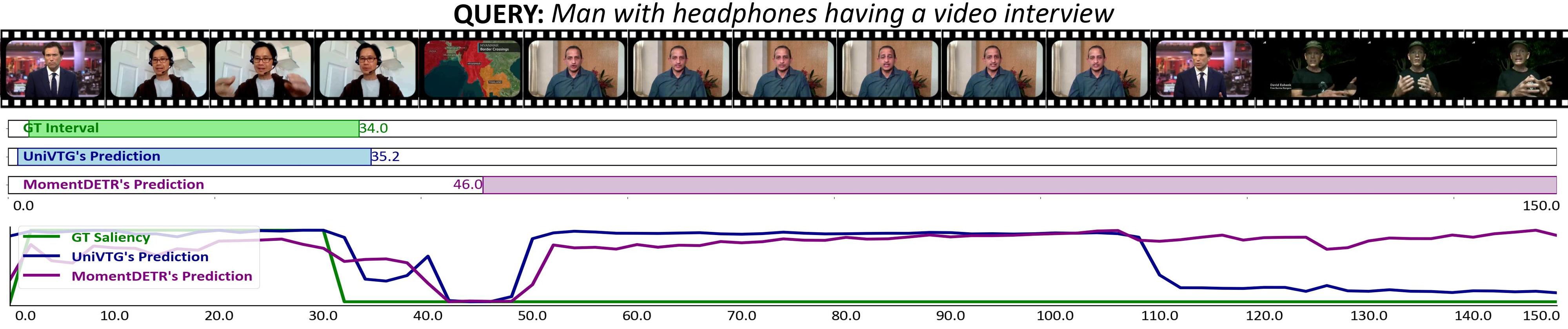}
    \end{minipage}

    \vspace{5pt}
    \begin{minipage}{\textwidth}
        \flushleft
        \captionsetup{labelformat=empty, justification=raggedright,singlelinecheck=false}
        \caption*{\large{\textbf{(b) \charades:} \textit{Indoor} domains, most videos are less than 1 minutes.}}
        \includegraphics[width=\textwidth]{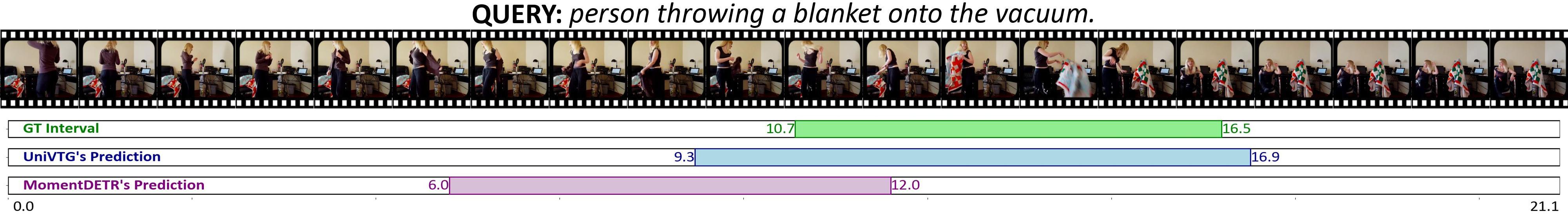}
    \end{minipage}

    \vspace{5pt}
    \begin{minipage}{\textwidth}
        \flushleft
        \captionsetup{labelformat=empty, justification=raggedright,singlelinecheck=false}
        \caption*{\large{\textbf{(c) \NLQ:} \textit{Egocentric} domain, videos are 8-20 minutes.}}
        \includegraphics[width=\textwidth]{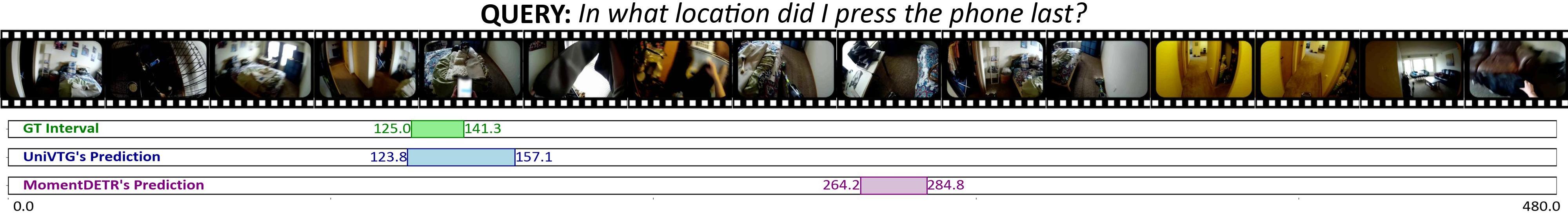}
    \end{minipage}

    \begin{minipage}{\textwidth}
        \flushleft
        \captionsetup{labelformat=empty, justification=raggedright,singlelinecheck=false}
        \includegraphics[width=\textwidth]{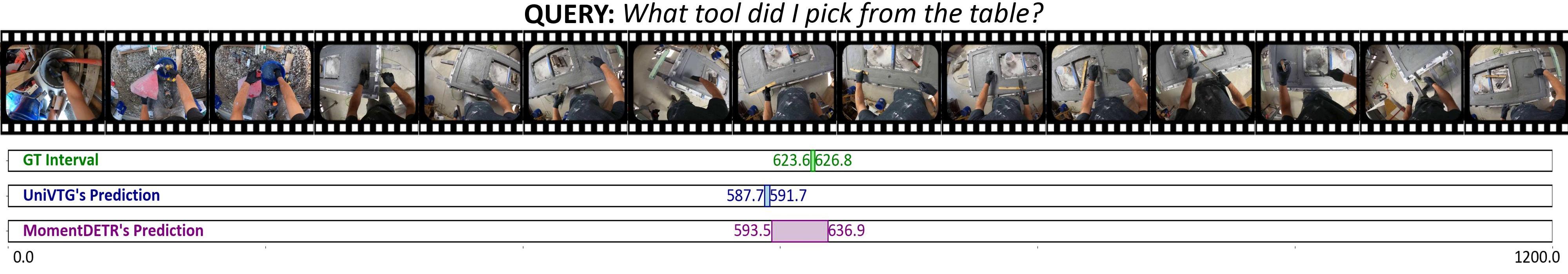}
    \end{minipage}

    \vspace{5pt}
    \begin{minipage}{\textwidth}
        \flushleft
        \captionsetup{labelformat=empty, justification=raggedright,singlelinecheck=false}
        \caption*{\large{\textbf{(d) \tacos:} \textit{Kitchen} domain, videos are average 4.8 minutes.}}
        \includegraphics[width=\textwidth]{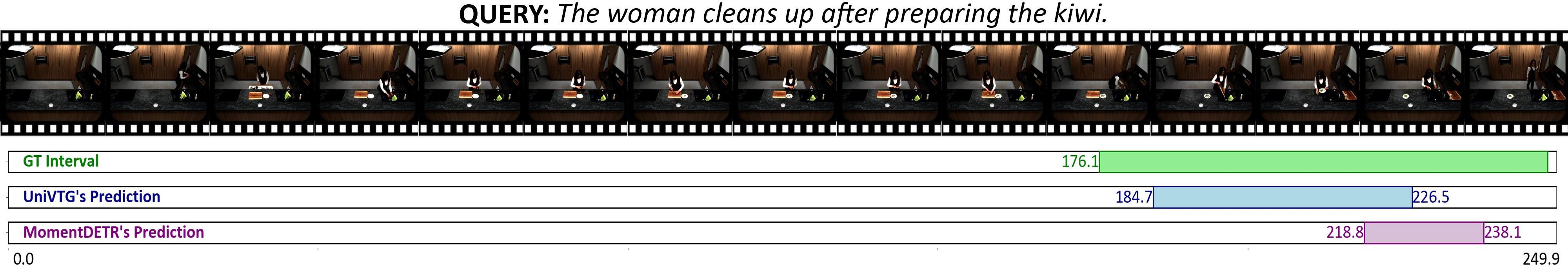}
    \end{minipage}
    \caption{Visualization of \textbf{Joint \mr~and \hl}~on (a)~\qv, and \textbf{\MR}~on (b)~\charades, (c)~\ego, (d)~\tacos. Textual queries are mostly \textit{natural sentences}.}
    \label{vis:1}
\end{figure*}

\begin{figure*}[ht]
    \begin{minipage}{\textwidth}
        \flushleft
        \captionsetup{labelformat=empty, justification=raggedright,singlelinecheck=false}
        \caption*{\large{\textbf{(e) \tvsum:} \textit{Web} diverse domain, videos are average 4.2 minutes long.}}
        \includegraphics[width=\textwidth]{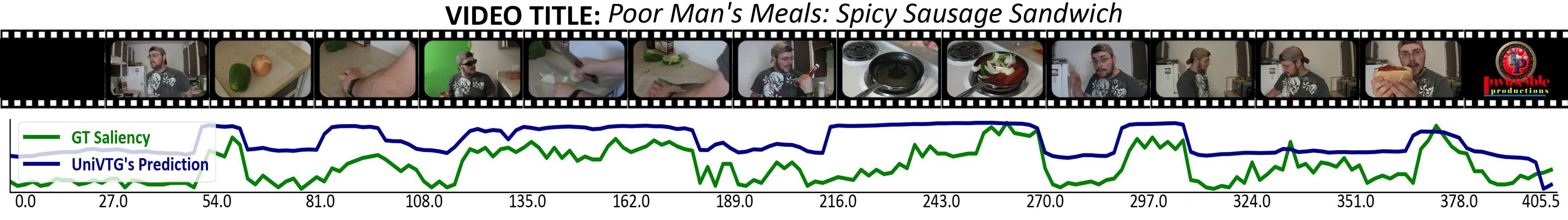}
    \end{minipage}

    \begin{minipage}{\textwidth}
        \includegraphics[width=\textwidth]{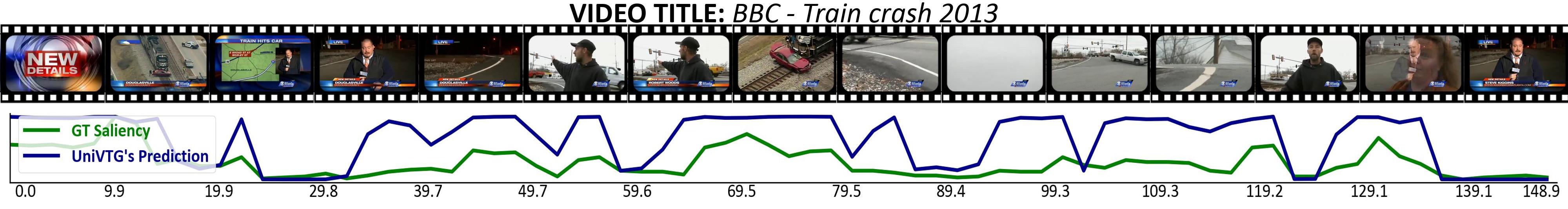}
    \end{minipage}

    \vspace{5pt}
    \begin{minipage}{\textwidth}
        \flushleft
        \captionsetup{labelformat=empty, justification=raggedright,singlelinecheck=false}
        \caption*{\large{\textbf{(f) \youtube:} \textit{Youtube} diverse domain, videos are average 1.5 minutes long.}}
        \includegraphics[width=\textwidth]{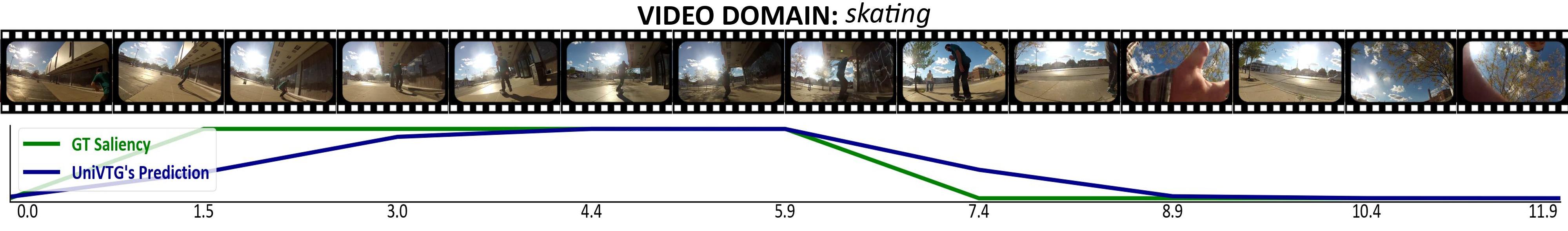}
    \end{minipage}

    \begin{minipage}{\textwidth}
        \includegraphics[width=\textwidth]{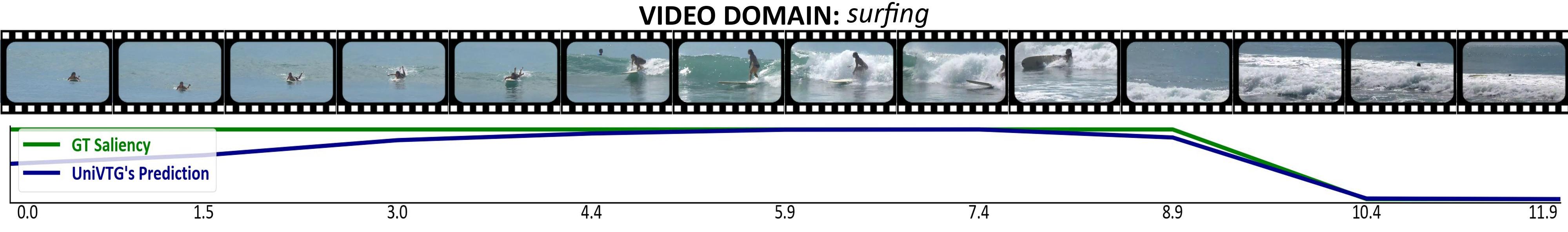}
    \end{minipage}

    \vspace{5pt}
    \begin{minipage}{\textwidth}
        \flushleft
        \captionsetup{labelformat=empty, justification=raggedright,singlelinecheck=false}
        \caption*{\large{\textbf{(g) Query-Focused Video Summarization:} \textit{Egocentric} domain, each video is between 3-5 hrs.}}        \includegraphics[width=\textwidth]{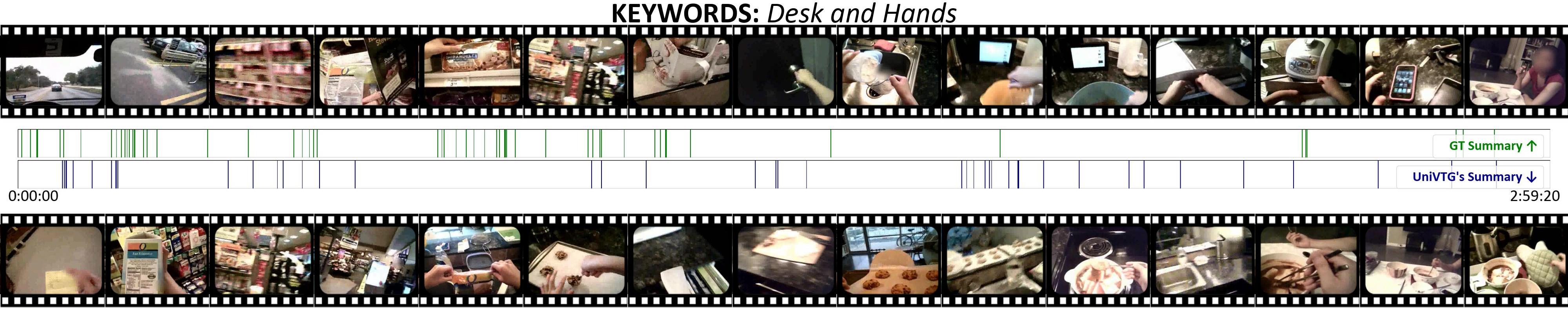}
    \end{minipage}

    \begin{minipage}{\textwidth}
        \centering
        \includegraphics[width=\textwidth]{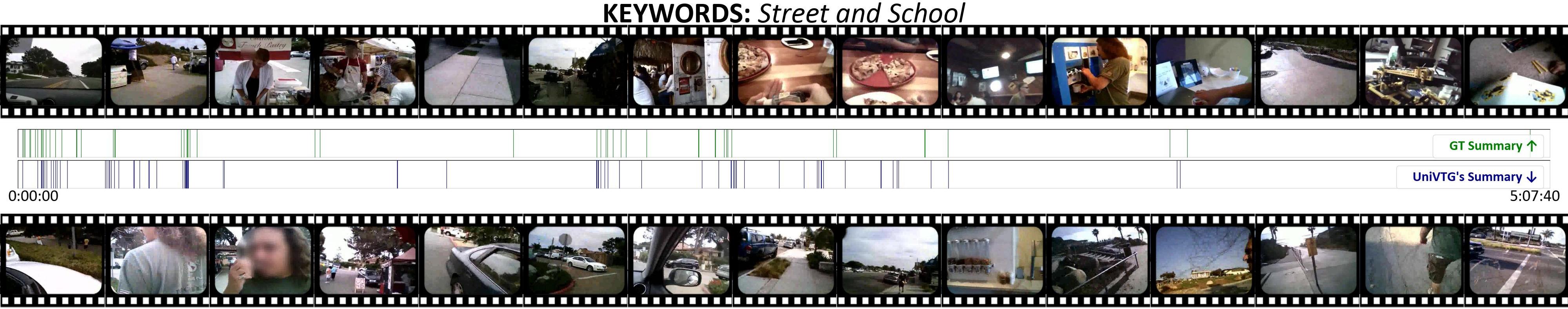}
    \end{minipage}
    \caption{Visualization of \textbf{\HL}~on (e)~\tvsum, (f)~\youtube; and \textbf{\VS}~on (g)~\qfvs. Textual queries can be \textit{video title}~(e), \textit{video domain}~(f), and \textit{keywords}~(g).}
    \label{vis:2}
\end{figure*}
\clearpage

\section{Acknowledgements}
This project is supported by the National Research Foundation, Singapore under its NRFF Award NRF-NRFF13-2021-0008, the DSO National Laboratories, Mike Zheng Shou's Start-Up Grant from NUS. The computational work for this article was partially performed on resources of the National Super computing Centre, Singapore.

{
\bibliographystyle{ieee_fullname}
\bibliography{main}
}

\clearpage
\section*{Appendix of \our}

\subsection*{A. CLIP teacher strategy}
The concept bank is a class list for open-world detection, sourced from here\footnote{\url{https://storage.googleapis.com/openimages/v6/oidv6-class-descriptions.csv}}. This list comprises $19,995$ class names, such as "Sandwich Cookies," "Air conditioning," and "Advertising." After conducting a manual check, we determined that the class list can effectively encompass the majority of common concepts.

In our approach, we begin by capturing frame-level clip image features from the video at a rate of 2 fps. Following this, we calculate their respective similarity scores in relation to the given class list. We then determine top-5 classes with the highest average scores, representing the most significant concepts within the video.

\begin{figure}[h]
\centering
\includegraphics[width=0.5\textwidth]{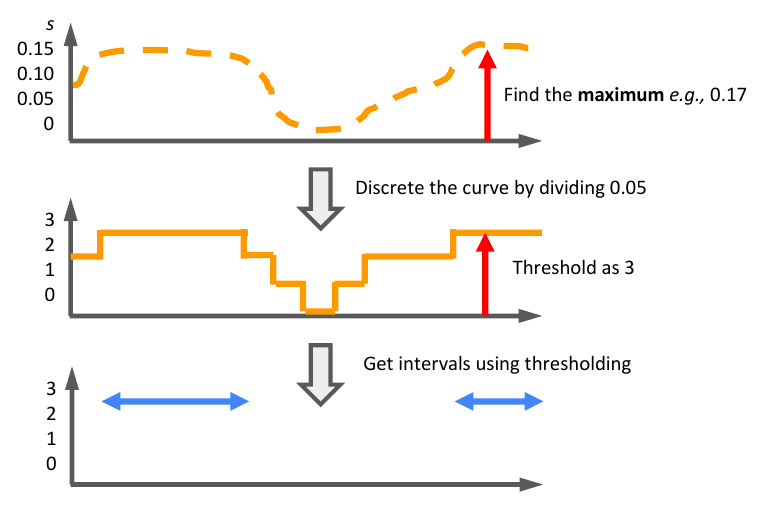}
\captionsetup{font={small}}
\caption{\small{Demonstration of how to threshold each video's curve.}}
\label{supp:fig:threshold}
\end{figure}

To derive intervals from the curve obtained from the diverse distributions, a fixed threshold is hard to determined and lacks the flexiblity. 
Thus, we discretize the continuous curve by a small value of $0.05$ and pick the maximum discrete value as our threshold. Then, adjacent clips that share the maximum discrete value to form an interval. In this way, we may produce multiple temporal windows from one video.  This process is shown in Fig.~\ref{supp:fig:threshold}.

\subsection*{B. Datasets}
\textbf{Pretraining corpus.}
To establish our pretraining corpus, we collect data through three ways:
For \point~labels, we extract the timestamped narrations from Ego4D~\cite{grauman2022ego4d} by \textit{excluding the NLQ val / test splits}.
For \interval~labels, we select a subset of videos~(less than 300K) sourced from VideoCC~\footnote{\url{https://github.com/google-research-datasets/videoCC-data}}, and treat their start and end timestamp as windows and caption as query.
For \curve~labels, we derive them from the above VideoCC subset videos.
Below, we describe the benchmarks used for the four settings separately.

\textbf{(i) Joint \MR~and \HL.}
{QVHighlights}~\cite{lei2021detecting} is the only dataset with available annotations for both \mr~and \hl, making it an ideal choice for benchmarking multi-task joint optimization.
This dataset contains $10,148$ videos with an average length of $150$ sec that covers daily vlogs, travel vlogs, and news events scenarios.
There are a total of $10,310$ queries associated with $18,367$ moments (on average, $1.8$ disjoint moments per query in the video). 

\textbf{(ii) \MR.}
We utilize three benchmarks to further evaluate \mr: \charades~\cite{gao2017tall}, \ego~\NLQ~(NLQ)~\cite{grauman2022ego4d} and \tacos~\cite{regneri2013grounding}.
{(a)}
\charades~contains $16,128$ indoor videos with an average length of $30.6$ sec, which are made up of $12,408$ query-interval pairs for training and $3,720$ query-interval pairs for testing.
{(b)}
NLQ focuses on daily egocentric scenarios, where videos are $8-20$ minutes long and queries are question, e.g.``What did i pour in the bowl?'', making this benchmark challenging. 
The training set contains $11.3$K annotated queries from $1$K videos, whereas the validation set contains $3.9$K queries from $0.3$K videos.
{(c)}
\tacos~contains $127$ videos with an average duration of $4.78$ minutes, where $75$ videos are used for training, $27$ and $25$ videos for validation and testing, respectively.

\textbf{(iii) \HL.}
We utilize two benchmarks to further evaluate \hl: \youtube~\cite{sun2014ranking} and \tvsum~\cite{song2015tvsum}.
{(a)}
\youtube~has $6$ domains with $433$ videos, where video titles are not provided, thus we use the domain name of each video as text queries.
{(b)}
While \tvsum~includes $10$ domains, each with $5$ videos, we use their video titles as text queries.
We follow \cite{liu2022umt} data splits that the ratio of training:testing is $0.8$:$0.2$. 

\textbf{(iv) \VS.}
We utilize the \qfvs~\cite{sharghi2017query} benchmark to evaluate the \vsum. This dataset includes the four videos in UT Egocentric dataset~\cite{lee2012discovering}. Each video is recorded in daily life and lasts between $3-5$ hours. Each query in this dataset is represented by two words from a total of $48$ pre-defined concepts.

\subsection*{C. Experimental settings}
\begin{table*}[t]
\centering
\footnotesize
\setlength{\tabcolsep}{2.5pt}
\begin{tabular}{llccccccccccccc}
\toprule
\textbf{Type} & \textbf{Datasets} & $l$  & {BS} & Epoch & Warmup & LR & Weight dacay & Gamma & LR drop & $\lambda_\text{SmoothL1}$ & $\lambda_\text{iou}$ & $\lambda_\text{f}$ & $\lambda_\text{intra}$ & $\lambda_\text{inter}$   \\
\midrule
Pretraining & $4.2$M corpus & $2$ & $64$ & $10$ & - &  $1e^{-4}$ & $1e^{-4}$ & - & - & $10$ & $1$ & $10$ & $0.1$ & $0.1$ \\
\midrule
\multirow{1}{*}{Joint MR \& HL} & \qv & $2$ & $32$ & $200$ & $10$ & $1e^{-4}$ & $1e^{-4}$ & $0.1$ & $80$ & $10$ & $1$ & $10$ & $0.05$ & $0.01$ \\
\midrule
\multirow{3}{*}{\MR} & \nlq & $2$ & $32$ & $200$ & $10$ & $1e^{-5}$ & $1e^{-5}$ & $0.1$ & $100$ &  $10$ & $1$ & $50$ & $0.1$ & $1.0$ \\
 & \charades & $1$ & $32$ & $100$ & $10$ & $1e^{-5}$ & $1e^{-5}$ & $0.1$ & $100$ & $10$ & $1$ & $10$ & $1.0$ & $0.5$ \\
 & \tacos & $2$ & $32$ & $100$ & $10$ & $1e^{-4}$ & $1e^{-4}$ & $0.1$ & $30$ & $10$ & $1$ & $10$ & $0.5$ & $0.1$ \\
\midrule
\multirow{2}{*}{\HL} & \youtube & $1^{\dagger}$ & $4$ & $100$ & $10$ & $1e^{-4}$  & $1e^{-4}$ & - &  - & $0$ & $0$ & $1$ & Search & $0$ \\
& \tvsum & $2$ & $4$ & $200$ & $10$ & $1e^{-4}$ & $1e^{-4}$ & - &  - & $0$ & $0$ & 1 & Search & $0$ \\
\midrule
\VS & \qfvs & $5$ & $20^{\ast}$ & $20$ & $0$ & $5e^{-5}$ & $5e^{-5}$ & - &  - & $0$ & $0$ & $1$ & $0.9$ & $0$ \\
\bottomrule
\end{tabular}
\captionsetup{font={small}}
\caption{\textbf{Parameter selections for each settings} 
where $l$ denotes the clip length; BS denotes the batch size; LR denotes the learning rate; LR drop denotes the learning rate drop up epoch; Warmup denotes the warmup epoch.
Search denotes to parameter searching individually for each domain.
$\dagger$ means \youtube~clips has overlapping frames, which is align with the \cite{liu2022umt}.
$\ast$ means batchsize in \qfvs~is based on the segment-level instead of  video-level.}
\label{supp:setting}
\end{table*}
(i) In Tab.~\ref{supp:setting}, we detail the parameters for each setting. Notably, for \hl~benchmarks \youtube~and \tvsum, which contain multiple domains treated as separate splits, we perform parameters tuning for $\lambda_\text{intra}$ within each domain. Then we aggregate the results obtained using optimal settings. The optimal settings are listed in Tab.~\ref{supp:youtube}-\ref{supp:tvsum}.

\begin{table}[h]
\centering
\footnotesize
\begin{tabular}{ccccccc}
\toprule
Domains & Dog & Gyn & Par. & Ska. & Ski. & Sur. \\
\midrule
$\lambda_\text{intra}$ & $0.6$ & $0.5$ & $0.4$ & $0.5$ & $0$ & $0.7$ \\
\bottomrule
\end{tabular}
\captionsetup{font={small}}
\caption{\small{Optimal $\lambda_\text{intra}$ under each domain in the Youtube HL.}}
\label{supp:youtube}
\end{table}

\begin{table}[h]
\centering
\footnotesize
\setlength{\tabcolsep}{3pt}
\begin{tabular}{ccccccccccc}
\toprule
Domains & BK & BT & DS & FM & GA & MS & PK & PR & VT & VU \\
\midrule
$\lambda_\text{intra}$ & $0.7$ & $0.9$ & $0.6$ & $0.4$ & $0.1$ & $0.1$ & $0$ & $0.6$ & $0.1$ & $0.5$ \\
\bottomrule
\end{tabular}
\captionsetup{font={small}}
\caption{\small{Optimal $\lambda_\text{intra}$ under each domain in the \tvsum.}}
\label{supp:tvsum}
\end{table}

(ii) During training, to maintain the balance between positive and negative samples, we allocate a weight of $0.1$ to the negatives ($f_i=0$) in binary cross-entropy loss Eq.~\ref{bce}.

(iii) When inferring highlights scores, we observe that $\{\tilde{f}_i+\tilde{s}_i\}_{i=1}^{L_v}$ can typically achieves better performance in \qv, while for smaller datasets \youtube~and \tvsum, using $\tilde{f}_i$ yield more reliable prediction.

(iv) For \vsum, we adhere to the same preprocessing settings in~\cite{xiao2020convolutional}, which extracts video frame features at $1$ FPS and take a $5$ seconds as a clip and compute the average frame feature within a clip to generate its clip-level feature. By applying the KTS algorithm~\cite{potapov2014category}, we split a long video into small segments under the conditions that the number
of segments in a video is no more than $20$ and each segment contains no more than $200$ clips.

\begin{table*}[!b]
\footnotesize
\centering
\setlength{\tabcolsep}{2pt}
\begin{tabular}{ccccc|ccccc|llll|ll|ll}
\toprule
\multicolumn{5}{c|}{\textbf{Pretraining}} & \multicolumn{5}{c|}{\textbf{Downstream}} & \multicolumn{2}{c}{\textbf{MR@QVHL}} & \multicolumn{2}{c|}{\textbf{HL@QVHL}} & \multicolumn{2}{c|}{\textbf{MR@NLQ}}  & \multicolumn{2}{c}{\textbf{MR@TaCoS}}  \\
$\mathcal{L}_\text{f}$ & $\mathcal{L}_\text{SmoothL1}$ & $\mathcal{L}_\text{iou}$ & $\mathcal{L}_\text{s}^\text{inter}$  & $\mathcal{L}_\text{s}^\text{intra}$  &
$\mathcal{L}_\text{f}$ & $\mathcal{L}_\text{SmoothL1}$ & $\mathcal{L}_\text{iou}$ & $\mathcal{L}_\text{s}^\text{inter}$  & $\mathcal{L}_\text{s}^\text{intra}$ & R$1$@$0.5$ & mAP & mAP & HIT@1 & R$1$@$0.3$ & mIoU & R$1$@$0.3$ & mIoU \\
\midrule
 &  &  &  &    &  \checkmark & \checkmark & & & & $54.71$ & $29.64$ & $33.12$ & $46.13$ & $5.96$ & $3.97$ & $48.46$ & $30.20$\\
 &  &  &  &    & \checkmark & \checkmark & \checkmark & & & $58.71$ & $35.89$ & $33.21$ & $45.03$ & $6.50$ & $4.43$ & $50.09$ & $32.42$\\
 &  &  &  &    &  \checkmark & \checkmark & \checkmark & \checkmark & & $59.16$ & $36.24$ & $38.59$ & $61.81$ & $6.97$ & $4.88$ & $51.14$ & $33.05$\\
 &  &  &  &    & \checkmark & \checkmark & \checkmark & \checkmark & \checkmark & $59.74$ & $36.13$ & $38.83$ & $61.81$ & $7.28$ & $4.91$ & $51.44$ & $33.60$\\
 \midrule
\checkmark &  &  &  &  & \checkmark & \checkmark & \checkmark & \checkmark & \checkmark & $62.00$ & $39.45$ & $39.59$ & $64.00$ & $8.83$ & $5.82$ & $52.04$ & $32.72$\\
\checkmark & \checkmark &  &  &  & \checkmark & \checkmark & \checkmark & \checkmark & \checkmark & $63.29$ & $40.43$ & $39.82$ & $64.19$ & $8.49$ & $5.73$ & $51.71$ & $34.76$\\
\checkmark & \checkmark & \checkmark &  & & \checkmark & \checkmark & \checkmark & \checkmark & \checkmark & $64.52$ & $41.65$ & $39.93$ & $63.68$ & $8.49$ & $5.74$ & $53.11$ & $34.48$\\
\checkmark & \checkmark & \checkmark & \checkmark &  &  \checkmark & \checkmark & \checkmark & \checkmark & \checkmark & $64.45$ & $41.84$ & $40.07$ & $64.32$ & $9.86$ & $6.52$ & $53.89$ & $36.76$\\
\checkmark & \checkmark & \checkmark & \checkmark &  \checkmark  & \checkmark & \checkmark & \checkmark & \checkmark & \checkmark & $68.39$ & $45.99$ & $41.25$ & $67.42$ & $11.74$ & $7.88$ & $56.11$ & $38.63$\\
 \bottomrule
\end{tabular}
    \captionsetup{font={small}}
 \caption{\small{\textbf{Ablation studies of downstream (top) and pretraining objective (bottom)} on \qv~{val} split, \nlq~val split and \tacos~val split.}
 }
\label{sup:aba_loss}
\end{table*}

During evaluation, we compute the foreground scores $\tilde{f}_i$ for each segment within a video, then aggregate these scores to derive an overall video score which is used to compute the metrics.
We calculate the conceptual similarity between each two video clip based on the intersection-over-union (IOU) of their related concepts. This conceptual similarity is then used as edge weights in a bipartite graph between two summaries, which aids in identifying the maximum weight match in the graph. Finally, precision, recall, and F1 scores can be determined based on the matching pairs.

\subsection*{D. Ablation studies of training objective}
Since we use identical training objectives during the stages of pretraining and downstream transferring. To gain a more thorough understanding of the impact each component has, we have constructed ablation studies as seen in Tab.~\ref{sup:aba_loss}, where the top half, we study the \textbf{effect of downstream training} objectives (without introduce any pretraining), while in the bottom half, we investigate the \textbf{effect of pretraining training} objectives (the downstream tuning use the same optimal parameter settings).


\subsection*{E. Parameters sensitivity}
\textbf{Transformer layers.}
In Tab.~\ref{sup:aba_layer}, we abalate the transformer layers $L\in [1,2,3,4,6,8]$ of multi-modal encoder in our unified model (without pretraining).

\begin{table}[!h]
\footnotesize
\centering
\setlength{\tabcolsep}{3.pt}
\begin{tabular}{c|cccc}
\toprule
\multirow{2}{*}{\textbf{\# Layers}} & \multicolumn{2}{c}{\textbf{MR}} & \multicolumn{2}{c}{\textbf{HD}}\\
&  R$1$@$0.5$ & mAP & mAP & HIT@1  \\
\midrule
$1$  & $47.16$& $26.62$ & $37.35$& $60.65$ \\
$2$  & $55.25$& $30.70$ & $38.33$& $60.52$ \\
$3$  & $59.03$& $34.06$ & $38.57$& ${62.13}$ \\
$4$  & ${59.74}$ & ${36.13}$ & ${38.83}$ & ${61.81}$\\
$6$  & ${61.55}$& ${39.88}$ & ${39.20}$& ${63.42}$ \\
$8$  & $60.32$& $38.24$ & $38.72$& $60.90$ \\
 \bottomrule
\end{tabular}
    \captionsetup{font={small}}
 \caption{\small{\textbf{Ablation studies of different transformer layers for multi-modal encoder} on \qv~{val} split.}
 }
 \centering
\label{sup:aba_layer}
\end{table}

\textbf{Projector dimension.}
In Fig.~\ref{sup:aba_proj}, we study the effect of projector dimension from $256$ to $1024$ (without pretraining).

\begin{figure}[!h]
  \centering
  \begin{minipage}[h]{0.48\linewidth}
  \centering
\begin{tikzpicture}
	\begin{axis} [
		axis x line*=bottom,
		axis y line*=left,
		legend pos=north east,
		ymin=38, ymax=48,
		xmin=256, xmax=1024,
		xtick={256,512,768,1024},
		ytick={38, 40, 42, 44, 46, 48},
   		ylabel={Avg. mAP},
            yticklabel pos=left,
            ylabel style={font=\Huge, yshift=-15pt},
		xlabel style={font=\Huge},
		width=\linewidth,
		legend style={cells={align=left}},
		label style={font=\footnotesize},
		tick label style={font=\footnotesize},
		legend style={at={(0.35,0.2)},anchor=west},
		]
		\addplot[mark=triangle,style={thick},NiceBlue] plot coordinates {
            (256, 43.18)
			(512, 44.25)
			(768, 45.61)
			(1024, 46.09)
		};
	\end{axis}
\end{tikzpicture}
  \captionsetup{font={scriptsize}, labelformat=empty}
    \vspace{-1pt}
    \caption*{\scriptsize{(a) Avg. mAP of \mr. \vspace{-1em}}}
  \end{minipage}
  \begin{minipage}[h]{0.48\linewidth}
  \centering
\begin{tikzpicture}
 	\begin{axis} [
		axis x line*=bottom,
		axis y line=left,
		legend pos=north east,
		ymin=60, ymax=68,
		xmin=256, xmax=1024,
		xtick={256,512,768,1024},
		ytick={60, 62, 64, 66, 68},
            ylabel={HIT@1},
            yticklabel pos=right,
            ylabel style={font=\Huge, yshift=-15pt},
		xlabel style={font=\Huge},
		width=\linewidth,
		legend style={cells={align=left}},
		label style={font=\footnotesize},
		tick label style={font=\footnotesize},
		legend style={at={(0.35,0.2)},anchor=west},
		]

		\addplot[mark=o,style={thick},orange] plot coordinates {
            (256, 64.81)
			(512, 66.21)
			(768, 66.63)
			(1024, 67.03)
		};
	\end{axis}
\end{tikzpicture}
  \captionsetup{font={scriptsize}, labelformat=empty}
  \vspace{-1pt}
    \caption*{\scriptsize{(b) HIT@$1$ of \hl.  \vspace{-1em}}}  
  \end{minipage}
    \captionsetup{font={small}}
  \caption{\small{\textbf{Ablation studies of projector dimension} on \qv~val split.}}
    \label{sup:aba_proj}
\end{figure}

\subsection*{F. Loss weights}
In Tab.~\ref{sup:aba_f}, we study the effect of foreground loss on three \mr~benchmarks (with pretraining).
\begin{table}[!h]
\footnotesize
\centering
\setlength{\tabcolsep}{3.0pt}
\begin{tabular}{c|cccccc}
\toprule
\multirow{2}{*}{\textbf{$\lambda_\text{f}$}} & \multicolumn{2}{c}{\textbf{\qv}} & \multicolumn{2}{c}{\textbf{NLQ}}  & \multicolumn{2}{c}{\textbf{\tacos}}\\
&  R$1$@$0.5$ & mAP &  R$1$@$0.3$ & mIoU  &  R$1$@$0.3$ & mIoU  \\
\midrule
$0.1$  & $66.97$& $46.02$ & $9.24$ & $6.64$ & ${46.51}$ & ${33.16}$\\
$0.5$  & $66.19$& $46.08$ & $9.50$& $6.75$  & ${50.21}$ & ${35.06}$\\
$1$  & $67.74$& ${46.22}$ & ${9.53}$& $6.80$ & ${51.79}$ & ${35.94}$\\
$5$  & ${67.35}$& $45.63$ & $9.89$& $6.88$ & ${54.01}$ & ${37.59}$\\
$10$ & ${67.81}$ &  $45.46$ & $7.26$ & $7.36$ & ${54.44}$ & ${37.55}$\\
$25$  & $68.00$& $45.06$ & $11.41$& $7.77$ & ${54.31}$ & ${37.27}$\\
$50$  & $66.71$& $44.32$ & ${11.13}$& ${7.49}$ & ${54.21}$ & ${35.61}$\\
 \bottomrule
\end{tabular}
    \captionsetup{font={small}}
 \caption{\small{\textbf{Ablation studies of foreground loss weight $\lambda_{\text{f}}$} on \qv, NLQ, and \tacos~\mr~benchmarks.}
 }
 \centering
\label{sup:aba_f}
\end{table}

\subsection*{G. Visualizations}
In Fig.~\ref{vis:1}~and \ref{vis:2}, we show quantitative visualizations of \our~predictions across different settings and domains.

\end{document}